\PassOptionsToPackage{table}{xcolor}
\PassOptionsToPackage{headheight=33pt}{geometry}
\documentclass[]{aidata}

\usepackage{amsmath,amsfonts,bm}

\def\eqref#1{equation~\ref{#1}}

\def\1{\bm{1}}

\DeclareMathAlphabet{\mathsfit}{\encodingdefault}{\sfdefault}{m}{sl}
\SetMathAlphabet{\mathsfit}{bold}{\encodingdefault}{\sfdefault}{bx}{n}

\usepackage{url}
\usepackage{multirow}
\usepackage{makecell}
\usepackage{booktabs}
\usepackage{colortbl}
\usepackage{array}
\usepackage{graphicx}
\usepackage{amsmath}
\usepackage{amssymb}
\usepackage{xspace}
\usepackage{caption}
\usepackage{placeins}
\usepackage{float}
\usepackage{wrapfig}
\usepackage{needspace}
\usepackage[ruled,linesnumbered]{algorithm2e}
\definecolor{newgreen}{RGB}{0,140,0}

\newlength{\cabinetfigurewidth}
\BeforeBeginEnvironment{wrapfigure}{%
    \Needspace{\dimexpr\cabinetfigurewidth+8\baselineskip\relax}%
}
\makeatletter
\AtBeginDocument{%
    \let\rfwa@original@nmfmt\NAT@nmfmt
    \renewcommand{\NAT@nmfmt}[1]{\mbox{\rfwa@original@nmfmt{#1}}}%
}
\makeatother

\newcommand{\maintablefont}{%
    \normalfont\fontsize{7}{9.5}\selectfont
    \renewcommand{\bffont}{\fontsize{7}{9.5}\selectfont}%
}
\newcommand{\maintablescale}{1.0}
\newcommand{\maintablebody}[1]{\scalebox{\maintablescale}{#1}}
\definecolor{maintablerow}{gray}{0.92}

\newcommand{\name}{\textsc{RFW-A}\xspace}

\makeatletter
\newif\if@appendixcontents
\newcommand{\appendixtableofcontents}{%
    \section*{Appendix Contents}%
    \begingroup
    \setcounter{tocdepth}{4}%
    \def\l@paragraph##1##2{\@dottedtocline{4}{5.5em}{4.5em}{##1}{##2}}%
    \@appendixcontentsfalse
    \let\appendix@savedcontentsline\contentsline
    \def\appendix@startanchor{appendix.A}%
    \def\contentsline##1##2##3##4{%
        \def\appendix@currentanchor{##4}%
        \if@appendixcontents
            \appendix@savedcontentsline{##1}{##2}{##3}{##4}%
        \else
            \ifx\appendix@currentanchor\appendix@startanchor
                \@appendixcontentstrue
            \fi
    \fi}%
    \@starttoc{toc}%
    \setcounter{tocdepth}{3}%
    \endgroup
}
\makeatother


\title{RFW-A: Agentic Procedural Generation of 400K Articulated Assets over 500 Categories}

\author[1,*]{Yuanbo Lyu}
\author[2,3,*]{Changjian Pan}
\author[2,4,*,\dagger]{Saining Zhang}
\author[1,*]{Yi Yang}
\author[4]{Peishuo Li}
\author[2]{Zhiheng Li}
\author[2]{Jialu Liu}
\author[2]{Junhao Chen}
\author[1]{Jiawei Weng}
\author[5]{Yupeng Gao}
\author[5]{Weiqing Xiao}
\author[1,2]{Wenyi Li}
\author[1,2]{Jiarui Guo}
\author[1,2]{Rui Zhang}
\author[1,2]{Jingrun Du}
\author[1]{Weiyi Lin}
\author[1,\ddagger]{Zenan Meng}
\author[1,\ddagger]{Hu Wei}
\author[1]{Lin Qu}
\author[2]{Zongzheng Zhang}
\author[2]{Hao Zhao}
\author[6,\ddagger]{Chaojian Li}

\affiliation[1]{Alibaba Group}
\affiliation[2]{Tsinghua University}
\affiliation[3]{Peking University}
\affiliation[4]{Nanyang Technological University}
\affiliation[5]{Nanjing University}
\affiliation[6]{The Hong Kong University of Science and Technology}

\contribution[*]{Equal contribution.}
\contribution[\dagger]{Project Leader.}
\contribution[\ddagger]{Corresponding authors.}
\checkdata[Email]{\{mengzenan.mzn, kongwang\}@alibaba-inc.com, chaojian@ust.hk}
\checkdata[Project Page]{\url{https://roboflywheel-com.github.io/RoboFlywheel-Articulation/}}

\abstract{
High-quality articulated assets are essential for embodied perception,
simulation, and manipulation, yet constructing them at scale remains difficult.
Manual annotation and procedural modeling require substantial human effort.
Recent agentic systems automate asset construction through program
synthesis, but repeat costly reasoning and refinement for every new asset.
Our key observation is that assets within a category can share construction
logic despite variations in structure and appearance. Agentic synthesis and
procedural generation are therefore complementary: agents can construct
reusable programs, while procedural execution reuses the resulting logic
across many samples.
We therefore present \name, which combines the two in a new articulated-asset
generation paradigm. Starting from existing articulated programs, \name
organizes reusable components and their dependencies into category-level
generators. Agents test and refine these generators across sampled
configurations and motion states.
Once constructed and validated, the generators produce diverse assets through
procedural execution with no further model calls or token charges. The
estimated API cost of generator authoring averages \$10.62, amortizing to
\$0.05 per asset over 200 assets, versus \$1.24 for the sampled per-asset
agentic synthesis baseline.
Using \name, we built \name-400K, containing over 400K sim-ready articulated assets
across more than 500 categories. Experiments show high structural and kinematic
validity and stable execution across physics simulators.
Beyond asset validity, training with \name-400K improves part-level articulation estimation on seen and unseen categories, while policies trained in simulation with retrieved and program-edited assets transfer zero-shot to real articulated objects.
}

\begin{document}
\raggedbottom
\maketitle

\begin{center}
    \centering
    \captionsetup{type=figure}
    \includegraphics[width=1.0\textwidth]{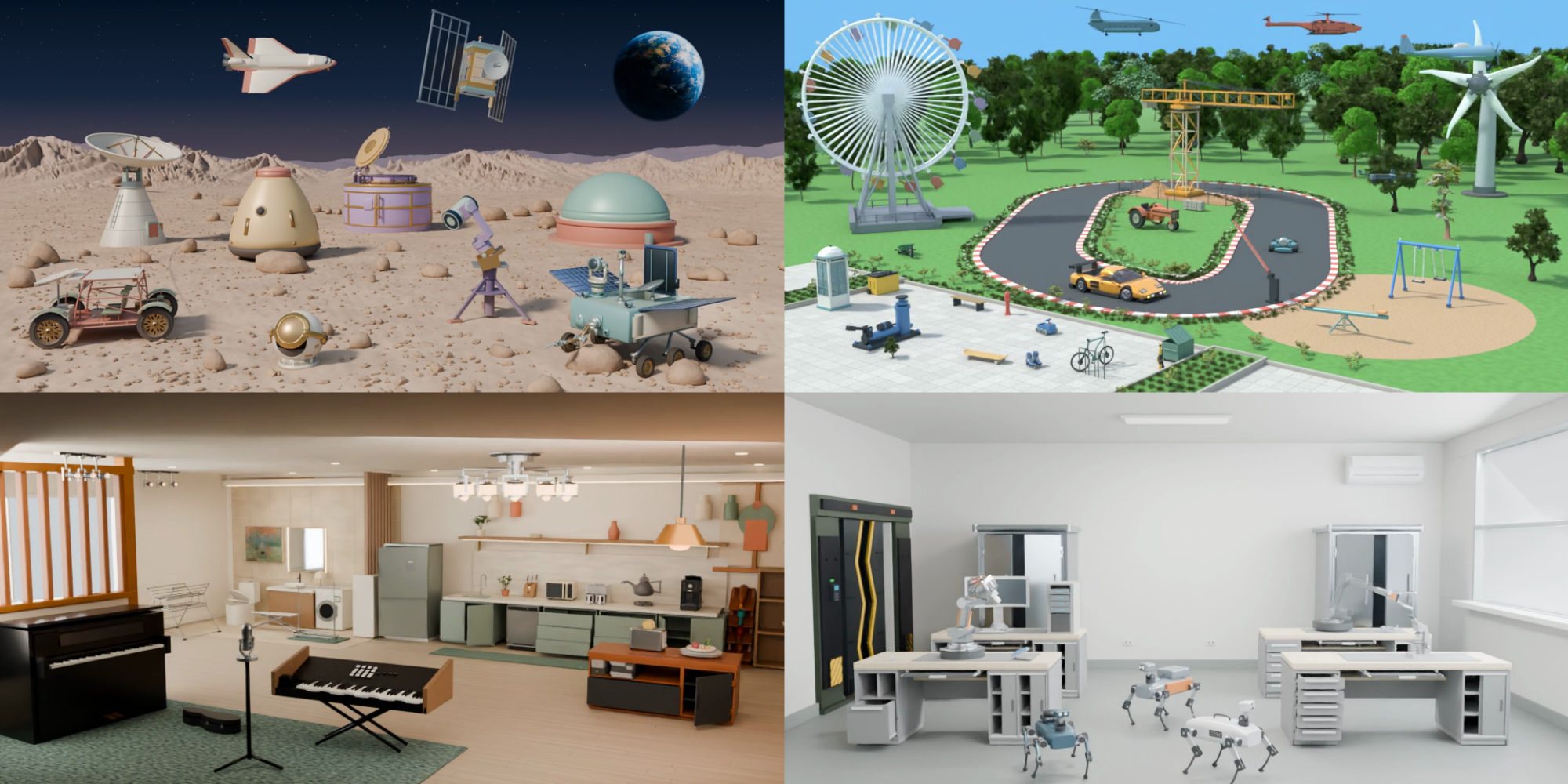}
    \caption{Overview of \name-400K, which provides 400K procedurally generated
    articulated assets spanning more than 500 object categories.}
    \label{fig:teaser}
\end{center}

\section{Introduction}

\begin{wrapfigure}{r}{\cabinetfigurewidth}
    \centering
    \vspace{-8pt}
    \includegraphics[width=1.0\linewidth]{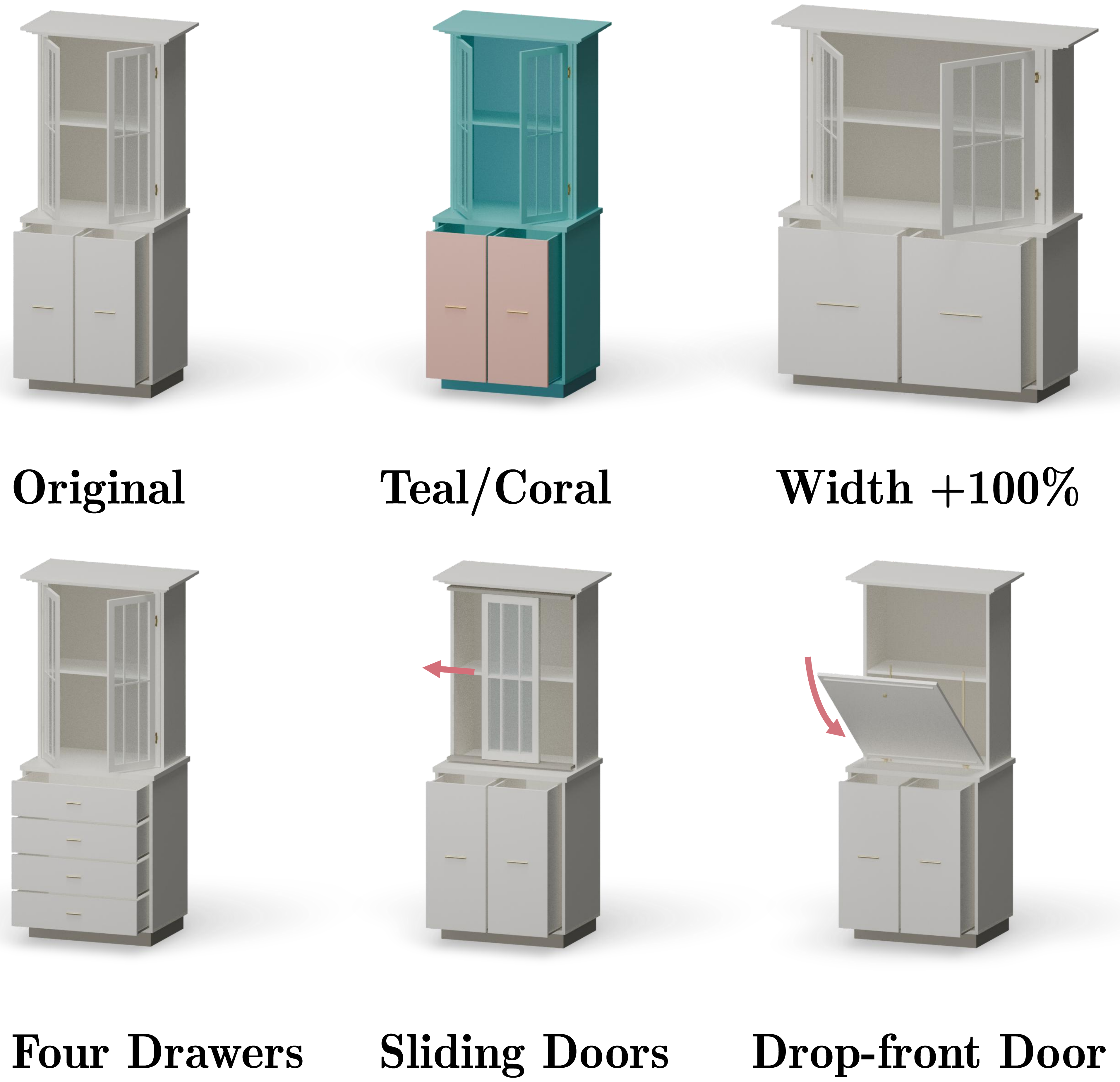}
    \caption{Cabinet families produced by reusable procedural generators.
    Across cabinet configurations, the generators vary materials, dimensions,
    door types, drawer layouts, and appearance while preserving shared rules
    for panel sizing, part placement, and joint attachment.}
    \label{fig:cabinet-generators}
    \vspace{-8pt}
\end{wrapfigure}

Articulated objects are ubiquitous in everyday environments, ranging from
cabinets and appliances to tools and mechanisms, as illustrated in
Figure~\ref{fig:teaser}. Their functionality depends not only on shape, but also on
how their parts are connected and move. High-quality articulated assets are
therefore essential for learning part semantics, kinematics, and interaction
affordances, as well as for training and evaluating manipulation policies in
simulation~\citep{sapien2020,partnet2019,akb482022,where2act2021,
vatmart2021,gamma2023,zhong2023implicittransporter}. Compared with static 3D
models~\citep{chang2015shapenet,objaverse2022,objaversexl2023}, these assets
require additional specifications of part hierarchies, joint parameters,
motion ranges, and physical properties. Producing and checking these
annotations requires substantial human effort~\citep{sapien2020,akb482022,
yang2026sketch2arti}; even with purpose-built interfaces, annotation can take
around ten minutes per asset~\citep{akb482022}.

Procedural generation reduces per-asset effort by encoding construction rules
in a reusable program that can produce many object variants
~\citep{infinitemobility2025}. Such generators can efficiently scale asset
sampling within an object category, but their construction rules are still
manually authored and are difficult to transfer across categories. Agentic
program synthesis addresses the authoring problem by using language models to
generate, execute, and refine articulated object programs
~\citep{articad2026,lam2026,articraft2026}. However, existing systems typically
synthesize assets one at a time, requiring a new round of reasoning and
correction for every output. Thus, procedural generation scales sampling but
not generator authoring, whereas agentic synthesis automates authoring but
repeats the process at the asset level.

These paradigms are complementary because assets within a category often share
construction logic. As shown in Figure~\ref{fig:cabinet-generators}, cabinets may vary in width,
door type, and drawer count while sharing rules for part dimensions,
placements, and joint attachments. Such logic can be encoded once in a
category-level generator and reused across many configurations;
Figure~\ref{fig:cabinet-generator-families} in the appendix extends this
example to four cabinet generator families. This motivates
our central question: \textit{can agentic synthesis automate the construction
of reusable category-level articulated object generators?}

\begin{figure}[!tbp]
    \centering
    \includegraphics[width=1.0\textwidth]{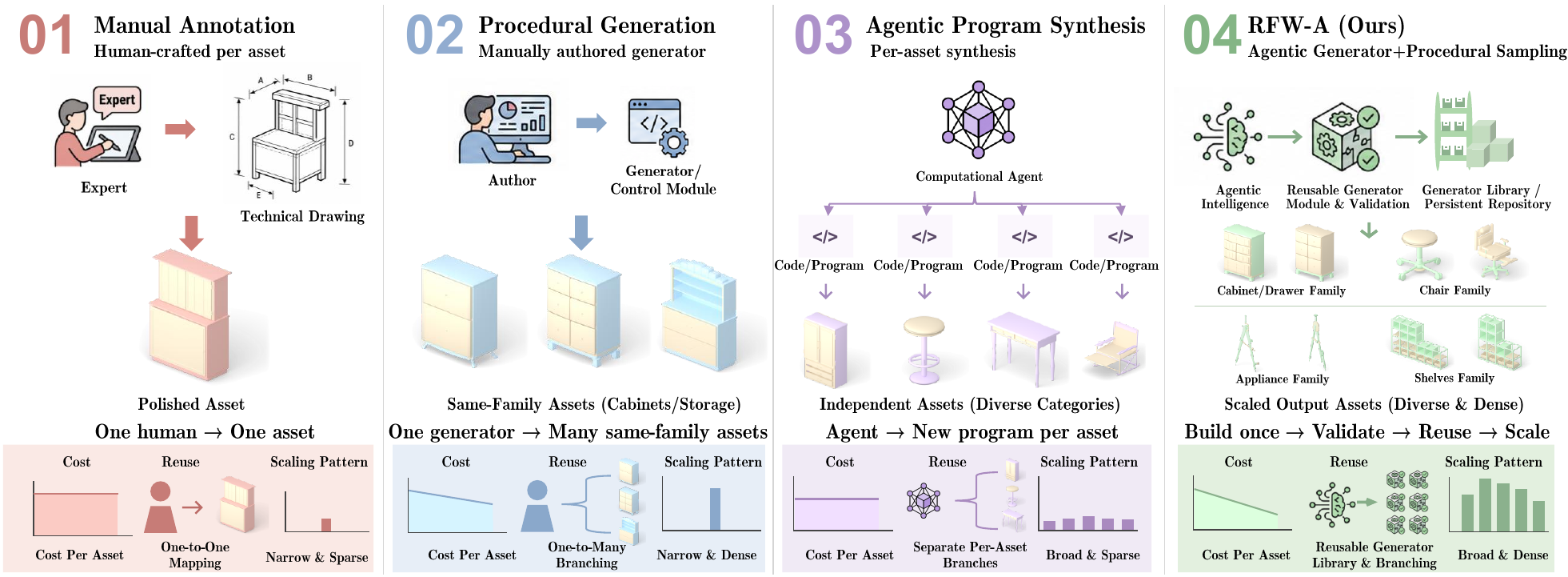}
    \caption{Comparison of articulated-asset construction paradigms.
    Manual annotation operates per asset; procedural generation amortizes
    sampling but requires category-specific generator authoring; and agentic
    synthesis repeats model inference for each asset. \name uses agentic
    synthesis to construct a reusable generator for each object family, then
    produces assets through procedural sampling.}
    \label{fig:paradigm}
\end{figure}

To answer this question, we present RoboFlywheel-Articulation (\name), an
agentic procedural generation framework that shifts agentic synthesis from
individual assets to reusable category-level generators
(Figure~\ref{fig:paradigm}). To support variation, each generator
specifies both what can change and which structural relationships must remain
valid. \name organizes these choices and dependencies into a category
blueprint built from reusable components in existing articulated programs.
Agents translate the blueprint into executable code, and iteratively test and
refine it across sampled configurations and motion states. Once constructed
and validated, the generator produces additional assets through procedural
execution with no further model calls or token charges. As summarized in Figure~\ref{fig:paradigm}, this
amortizes agentic construction cost across many assets and enables broad and
dense scaling across object categories: in our cost comparison, amortizing
the mean generator-authoring API cost over 200 assets yields \$0.05 per asset,
versus \$1.24 for the sampled per-asset agentic synthesis baseline
(Figure~\ref{fig:cost-scaling}).

Using \name, we build \name-400K, containing 402{,}410 articulated assets
across more than 500 object categories, approximately 37 times the size of
Articraft, the largest existing articulated-object dataset. Each asset
includes part structure, kinematics, estimated physical properties, and the
parameters needed to reproduce its construction.
Our evaluation shows a 100\% asset-level kinematic pass rate over the full
release. On a 200-asset evaluation subset, 99.5\% of assets pass a 1-second
passive rollout with the root fixed in all four simulators: Genesis,
PyBullet, MuJoCo, and Isaac Sim.
Downstream experiments further show that training with \name-400K improves
part-level articulation estimation on seen and unseen categories compared
with PartNet-Mobility alone and Articraft augmentation.
We further evaluate its utility for robot learning: across five manipulation
tasks, policies trained from simulated demonstrations using retrieved and
program-edited assets transfer zero-shot to real articulated objects.

\Needspace{6\baselineskip}
In summary, our contributions are:
\begin{itemize}
    \item We introduce \name, an agentic procedural generation framework
    that constructs and validates reusable category-level generators,
    enabling asset sampling without per-asset model inference.

    \item Using \name, we release \name-400K, comprising 402{,}410 articulated
    assets across more than 500 object categories. Each asset includes explicit
    part structure, kinematics, estimated physical properties, and reproducible
    generation parameters.

    \item We evaluate asset quality and cross-simulator compatibility, demonstrate improved part-level articulation estimation on seen and unseen categories, and show zero-shot sim-to-real manipulation using policies trained with retrieved and program-edited \name-400K assets.
\end{itemize}

\section{Related Work}

\subsection{Articulated Object Datasets}

Large-scale 3D repositories provide broad geometric diversity, but most assets
are static and lack reliable part and articulation annotations
~\citep{chang2015shapenet,objaverse2022,objaversexl2023}. Datasets for
articulated objects address this gap by adding part hierarchies, joint
parameters, motion ranges, and actionable part labels to curated CAD models or
real scans~\citep{shape2motion2019,sapien2020,akb482022,gapartnet2022}. More
recent releases expand category and structural coverage, and some also include
material or physical-property annotations
~\citep{yang2026sketch2arti,artiverse2026,physxanything2026}. Complementing
these annotated collections, \name-400K provides both assets and their
executable construction programs. These programs support resampling and
editing within each defined object family.

\subsection{Articulated Object Reconstruction and Generation}

Articulated reconstruction methods recover instance geometry and kinematics
from interaction, single- or multi-state images, RGB-D observations, or videos
~\citep{ditto2022,paris2023,digitaltwinart2024,real2code2025}. Recent methods
use 3D Gaussian Splatting to improve reconstructed geometry and appearance
~\citep{artgs2025,reartgs2025,reartgspp2026,videoartgs2025,freeartgs2026,shen2026gaussianart,huang2026arthoi,gao2025partrm}.
These approaches can produce detailed digital twins, but they reconstruct a
particular observed object and must infer occluded geometry and ambiguous motion
from the available observations.

Generative methods instead learn distributions over part geometry,
connectivity, and kinematics using graph-based, diffusion, autoregressive, or
structured latent representations
~\citep{nap2023,cage2024,meshart2025,artilatent2025,uniart2025,he2026spark}.
Image-, text-, and sketch-conditioned methods provide additional control over
articulated asset modeling~\citep{singapo2025,dreamart2025,artllm2026,
yang2026sketch2arti}. Another line of work adds articulation to existing
geometry or assembles articulated assets
from observed or retrieved components
~\citep{urdformer2024,urdfanything2025,articulateanything2024,
articulateanymesh2025,particulate2026,instructparticulate2026,
motionanymesh2026}. Related work on deformable assets considers skeleton
and skinning generation~\citep{drive2025,animatorskeleton2026}.
\name instead constructs explicit category-level programs whose structural
choices and parameters can be varied directly to generate new rigid-link
object instances.

\subsection{Programmatic, Procedural, and Agentic Generation}

Programmatic representations encode 3D construction as CSG trees, assembly
programs, or CAD command sequences. Their explicit operations and parameters
support structured generation and editing
~\citep{csgnet2018,shapeassembly2020,shapemod2021,deepcad2021,text2cad2024,
cadcrafter2025,cadllama2025}. Procedural generation reuses such programs
across sampled configurations~\citep{xu2025diffusionknowstransparency}.
Arti-PG synthesizes articulated objects by manipulating structure programs
and propagating annotations~\citep{artipg2025}, while Infinigen-Articulated
and Infinite Mobility provide manually designed category-level
generators~\citep{joshi2025infinigenarticulated,infinitemobility2025}.

LLMs and agents have also been used for code generation and structured 3D or
motion synthesis through programmatic interfaces, tool use, and intermediate
representations
~\citep{threedgpt2023,scenecraft2024,ll3m2025,idea23d2025,paircoder2026,
smoogpt2025}. For static shapes, ShapeLib uses LLMs to construct reusable
procedural abstractions and refine a shape-sampling
program~\citep{shapelib2025}. Recent systems specialize agentic synthesis to
articulated assets by jointly constructing geometry and kinematics as
executable programs
~\citep{lam2026,articad2026,articraft2026}. These systems broaden the range of
objects that can be synthesized, but still run generation and refinement for
each asset. \name instead uses agents to construct category-level generators
from existing articulated programs and refine them using configuration and
motion tests. The resulting generators produce new assets through procedural
execution without additional model inference.

\section{Method}
\label{sec:method}

\begin{figure}[t]
    \centering
    \includegraphics[width=1.0\textwidth]{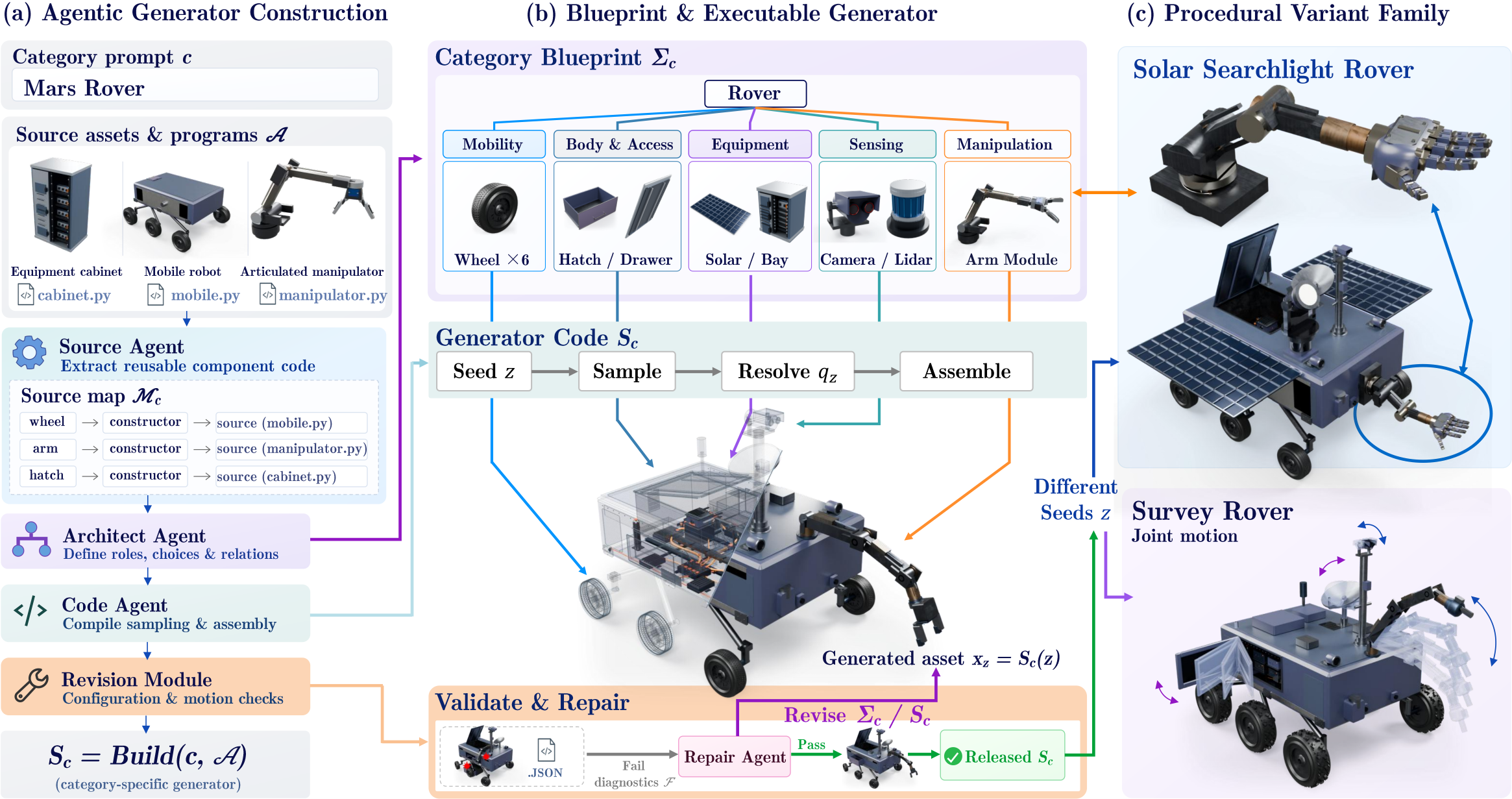}
    \caption{\textbf{Overview of the \name generator-construction pipeline.}
    Panel (a) shows how agents extract reusable components from source assets
    and executable programs, organize them into a category blueprint, and
    compile the blueprint into a generator. Panel (b) shows procedural
    execution by sampling choices, resolving dependent parameters, and
    assembling an asset, followed by validation and repair. Panel (c) shows
    that a released generator produces a family of articulated variants from
    different seeds, while exposing the joint motion encoded by the generator.}
    \label{fig:pipeline}
\end{figure}

\name combines agentic program synthesis with procedural generation,
as shown in Figure~\ref{fig:pipeline}. Given a category prompt $c$ and
source assets with executable CAD or 3D modeling programs
$\mathcal{A}$~\citep{articraft2026}, \name constructs a reusable
category-level generator $S_c$. Four agents implement this workflow:
the Source Agent extracts reusable component code, the Architect Agent
constructs the category blueprint, the Code Agent implements the
executable generator, and the Repair Agent revises the blueprint or
generator using structured diagnostics. The Revision Module coordinates
configuration and motion checks. We write construction and sampling as
\[
    S_c=\operatorname{Build}(c,\mathcal{A}),\qquad x_z=S_c(z),
\]
where $\operatorname{Build}$ includes construction, validation, and
release, and $z$ is a random seed. After release, different seeds produce
different assets through procedural sampling without further model
inference. Algorithm~\ref{alg:pva} summarizes the complete loop.

\subsection{Agentic Generator Construction}
\label{sec:source_grounding}
\label{sec:generator_architecture}

To construct $S_c$, we first extract reusable component code from
$\mathcal{A}$. We then build a category blueprint $\Sigma_c$ to specify
the available components, adjustable parameters, and assembly and joint
rules. This blueprint guides code generation.

\paragraph{Source component extraction.}
The Source Agent identifies relevant components in $\mathcal{A}$ based on
the category prompt $c$. It extracts their construction code into a source
map $\mathcal{M}_c$:
\[
    \mathcal{M}_c = \operatorname{Extract}(c,\mathcal{A}).
\]
The map links each component to its construction code and source program.
In Figure~\ref{fig:pipeline}, the rover reuses wheel code from the mobile-robot
program and arm code from the manipulator program.

\paragraph{Blueprint construction.}
The Architect Agent constructs $\Sigma_c$ from the components in
$\mathcal{M}_c$. We represent the blueprint as a graph $G_c=(V_c,E_c)$.
Each node specifies a component role, its available designs, and adjustable
parameters. It also indicates whether the component is optional or repeated.
Each edge specifies how two components attach, including joint parameters
for movable connections.
The blueprint also records rules for computing dependent dimensions and
placements when component choices or parameters change. These construction
rules are shared across configurations.

\subsection{Procedural Generator and Variant Sampling}
\label{sec:generator_coding}

The Code Agent implements the blueprint $\Sigma_c$ as an executable
generator $S_c$, using the component code in $\mathcal{M}_c$.
Given a seed $z$, the generator selects components and samples free
parameters. It computes dependent dimensions, placements, and joint parameters
from the blueprint's rules. The program then calls the component construction
code to build and assemble the asset. The three execution steps in
Figure~\ref{fig:pipeline} are
\[
    \chi_z = \operatorname{Sample}(\Sigma_c;z),\qquad
    q_z = \operatorname{Resolve}(\Sigma_c,\chi_z),
    \qquad
    x_z = S_c(z) = \operatorname{Assemble}(\mathcal{M}_c,q_z).
\]
Here $\chi_z$ contains the sampled choices and free parameters, while $q_z$
contains the selected components and all resolved construction parameters.
These steps execute the rules encoded in $S_c$ without further model
inference. The program exports meshes, rigid links, collision geometry,
joints, and a URDF file. It also records the source of each reused component.

\subsection{Validating, Repairing, and Releasing the Generator}
\label{sec:validation}
\label{sec:diagnostic_repair}

A generator that produces one valid asset may fail under other configurations
or joint states. Before large-scale sampling, the Revision Module therefore
checks blueprint--code consistency and tests assets across component choices,
parameter settings, and joint states.

\paragraph{Blueprint--code consistency.}
We check that each choice in $\Sigma_c$ has a code path and that its
dependent parameters can be computed. We run sampling and parameter resolution
across seeds until each choice has been selected multiple times. These checks
detect missing branches, unused parameters, and incomplete dependencies
without constructing geometry.

\paragraph{Configuration and motion testing.}
We test progressively larger batches generated with deterministic seeds.
Each batch must pass before we increase its size. Additional tests target
uncovered component choices, parameter boundaries, and selected component
combinations. Each asset is checked for successful geometry construction, part
connectivity, URDF export and parsing, and valid articulation. Collision checks
cover the rest pose, multiple states across each joint's range, and selected
coupled-joint states.

Let $\mathcal{Z}_{\mathrm{test}}$ contain the current validation cases and all
retained regression cases, and let $\mathcal{V}_c$ collect the component
choices declared in $\Sigma_c$. The generator is released only when every
retained case passes and every declared choice is covered by a passing case:
\[
    \operatorname{Accept}(S_c)
    \;\Longleftrightarrow\;
    \left[\forall z\in\mathcal{Z}_{\mathrm{test}},\;
    \operatorname{Valid}(x_z)\right]
    \;\wedge\;
    \left[\forall v\in\mathcal{V}_c,\;\exists z\in\mathcal{Z}_{\mathrm{test}}:
    v\in q_z\right],
\]
where $\operatorname{Valid}(x_z)$ denotes that $x_z$ passes the checks above.

\paragraph{Failure localization and repair.}
For each failure, we record the failed check, sampled configuration, affected
parts or joints, and responsible construction step. The report also includes
measurements such as penetration depth. The Repair Agent uses these diagnostics
to update the blueprint or generator, or both:
\[
    \bigl(\Sigma_c^{(t+1)},S_c^{(t+1)}\bigr)
    = \operatorname{Repair}\left(
        \Sigma_c^{(t)},S_c^{(t)},\mathcal{F}_t
    \right),
\]
where $\mathcal{F}_t$ is the structured diagnostics collected at iteration
$t$. Blueprint errors are returned to the Architect Agent, and blueprint
repairs are followed by recompilation of the generator. Implementation errors
go to the Code Agent with the blueprint unchanged. The Revision Module then
retests both failed and previously passed cases, retaining failures as
regression tests until the release criterion is met.

\providecommand{\stackcell}[1]{\shortstack{#1}}
\providecommand{\best}[1]{\ensuremath{\boldsymbol{#1}}}

\section{\name-400K Dataset: Articulated Assets}
\label{sec:dataset}

\name{} turns the validated generators of Section~\ref{sec:validation}
into a dataset by sampling a fixed seed budget for each generator. The
release retains both the exported assets and their construction records,
linking each sample to the program and configuration that produced it.

\subsection{Dataset Overview}
\label{sec:dataset_overview}

The \name-400K release contains 402{,}410 articulated assets covering more than
500 semantic object categories.
Alongside visual and collision geometry, each asset retains its selected
components, resolved parameters, part hierarchy, joint parameters, and source
provenance. Its generator version and seed identify a reproducible sample,
while the recorded configuration exposes the choices used to construct it.

\begin{table}[!htbp]
\centering
\caption{
Release-level comparison of articulated-object datasets, their construction
types, and available annotations. Cats. denotes categories, Proc. denotes
procedural generation, and Metric denotes metric-scale geometry.
\textbf{Physical Grounding} groups the
physical-property fields (Section~\ref{sec:dataset-material-editing}): $E$ is
Young's modulus, $\nu$ is Poisson's ratio, Friction includes static and
dynamic coefficients, and $e$ is the coefficient of restitution.
\textbf{F-mat.} marks fine-grained material annotation, i.e., material
assignments below the articulated-part level. $\checkmark$ indicates that a field or resource is included in the
released data or documented format, $\triangle$ indicates partial or
format-dependent availability, and $\times$ indicates that it is absent. The
RFW-A-400K row follows the release specification and evaluation manifest.
}
\label{tab:table0}

\maintablefont
\setlength{\tabcolsep}{2.2pt}
\renewcommand{\arraystretch}{1.08}
\maintablebody{%
\begin{tabular}{@{}l@{\hspace{4pt}}r@{\hspace{4pt}}r@{\hspace{4pt}}l%
*{8}{>{\centering\arraybackslash}m{24pt}}@{}}
\toprule
\multirow{2}{*}{\textbf{Dataset}} &
\multirow{2}{*}{\textbf{Assets}} &
\multirow{2}{*}{\textbf{Cats.}} &
\multirow{2}{*}{\textbf{Type}} &
& & &
\multicolumn{5}{c}{\textbf{Physical Grounding}} \\
\cmidrule(lr){8-12}
& & & &
\makebox[\linewidth]{\textbf{Metric}} &
\makebox[\linewidth]{\textbf{Texture}} &
\makebox[\linewidth]{\textbf{F-mat.}} &
\makebox[\linewidth]{\textbf{Density}} &
$\boldsymbol{E}$ &
$\boldsymbol{\nu}$ &
\makebox[\linewidth]{\textbf{Friction}} &
$\boldsymbol{e}$ \\
\midrule

PartNet-Mobility & 2,347 & 46 & Manual & $\times$ & $\triangle$ & $\times$ &
$\times$ & $\times$ & $\times$ & $\times$ & $\times$ \\
Infinite Mobility & On-demand & 22 & Procedural & $\times$ & $\triangle$ &
$\times$ & $\times$ & $\times$ & $\times$ & $\times$ & $\times$ \\
\mbox{Infinigen-Articulated} & 8,226 & 17 & Procedural & $\triangle$ &
$\checkmark$ & $\times$ & $\triangle$ & $\times$ & $\times$ & $\triangle$ &
$\times$ \\
PhysX-Mobility & 2,024 & 45 & Manual & $\times$ & $\triangle$ & $\times$ &
$\checkmark$ & $\checkmark$ & $\checkmark$ & $\times$ & $\times$ \\
Artiverse & 3,544 & 84 & Manual & $\times$ & $\checkmark$ & $\times$ &
$\checkmark$ & $\times$ & $\times$ & $\times$ & $\times$ \\
LAM & 3,217 & 650 & Agentic & $\times$ & $\triangle$ & $\times$ & $\times$ &
$\times$ & $\times$ & $\times$ & $\times$ \\
SketchMobility & 4,956 & 47 & Manual & $\times$ & $\triangle$ & $\times$ &
$\times$ & $\times$ & $\times$ & $\times$ & $\times$ \\
Articraft & 10,787 & 244 & Agentic & $\times$ & $\times$ & $\times$ & $\times$ &
$\times$ & $\times$ & $\times$ & $\times$ \\
\midrule
\rowcolor{maintablerow}
\textbf{RFW-A-400K (Ours)} & \textbf{402,410} & \textbf{531} &
\textbf{Agentic + Proc.} & $\mathbf{\checkmark}$ & $\mathbf{\checkmark}$ &
$\mathbf{\checkmark}$ & $\mathbf{\checkmark}$ & $\mathbf{\checkmark}$ &
$\mathbf{\checkmark}$ & $\mathbf{\checkmark}$ & $\mathbf{\checkmark}$ \\
\bottomrule
\end{tabular}}

\end{table}

\begin{figure}[!h]
\centering
\includegraphics[width=1.0\textwidth]{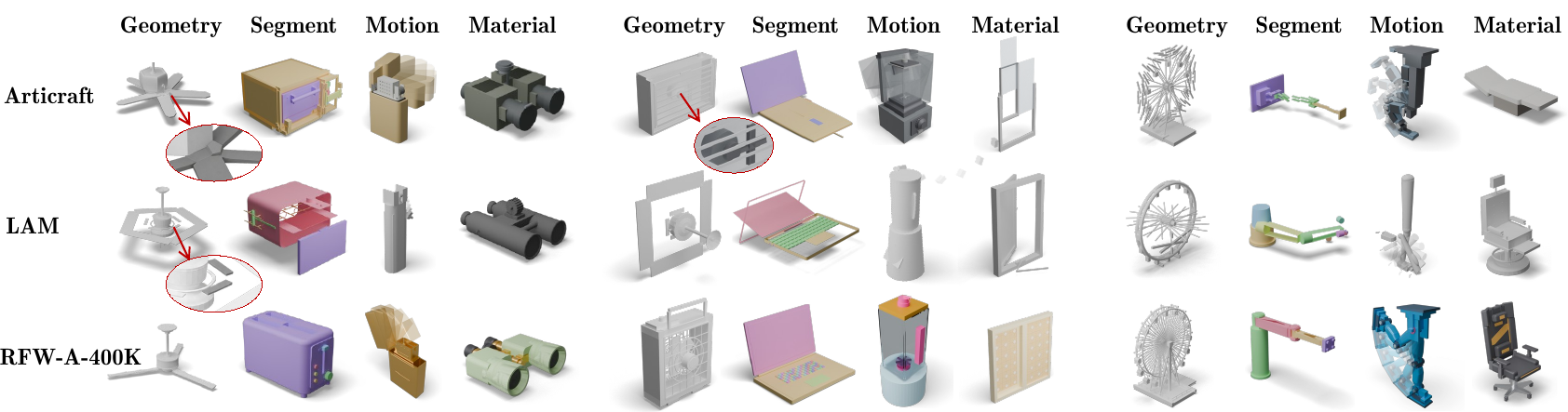}
\caption{Qualitative comparison of Articraft, LAM, and \name-400K.
The three rows are the compared datasets; each row contains four quality
 dimensions (geometry, part segmentation, motion, and material appearance).}
\label{fig:qualitative-comparison}
\end{figure}

Compared with existing articulated-object datasets, \name-400K combines this
scale with material assignments below the articulated-part level and
associated physical-property annotations (Table~\ref{tab:table0}). Each asset
is represented through geometry, part structure, joint motion, and surface
materials, which provide complementary views of asset quality in the
qualitative comparison with Articraft and LAM
(Figure~\ref{fig:qualitative-comparison}).

\begin{figure}[t]
\centering
\includegraphics[width=1.0\textwidth]{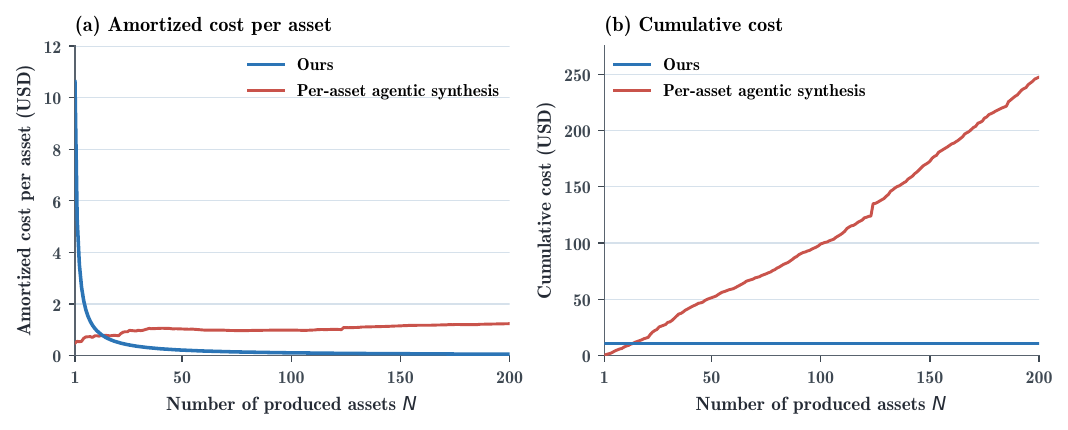}
\caption{API-cost scaling of reusable category-level generators versus per-asset
agentic synthesis. (a) Our mean generator-authoring cost is amortized over $N$
assets, while the baseline averages $N$ recorded synthesis costs. (b) Our
one-time API cost stays fixed, while the baseline accumulates a new charge
for each asset.}
\label{fig:cost-scaling}
\end{figure}

\name{} pays the agentic authoring cost once per generator, then produces
additional assets procedurally without further model calls or token charges.
Amortizing the mean generator-authoring cost of \$10.62 over 200 assets yields
an API cost of \$0.053 per asset, versus \$1.237 for the sampled per-asset
agentic synthesis baseline, a 23-fold cost difference
(Figure~\ref{fig:cost-scaling}). We estimate these costs using ten
category-level generators and 200 successful Articraft synthesis
records~\citep{articraft2026}, twenty from each of the same categories. 

\subsection{Material Grounding and Directed Editing}
\label{sec:dataset-material-editing}

The procedural representation specifies how an asset is built and moves;
material grounding extends it to the appearance and physical properties of
its surfaces. To apply this grounding across the release, each validated
generator records a material-class identifier for each visual surface and
connects it to a shared library built once offline. The library supplies
concrete instances that can be reused across generators and sampled across
assets. After geometry and articulation validation, a single offline pass
assigns each visual surface a class from the controlled taxonomy; performing
this pass after repair ensures that added, removed, or renamed surfaces receive
the correct identifiers.

The library contains 8{,}295 curated PBR materials organized into 124 leaf
classes across 18 families. Each entry associates a PBR appearance with six
physical properties in SI units: density, Young's modulus, Poisson's ratio,
static and dynamic friction coefficients, and coefficient of restitution.
These properties are estimated offline under fixed reference conditions;
their uncertainty intervals and confidence estimates are retained in the
library (Appendix~\ref{app:material-physics}). Keeping appearance and physical
annotations in the same entry lets both follow the material selected for a
surface.

\begin{figure}[t]
\centering
\includegraphics[width=1.0\textwidth]{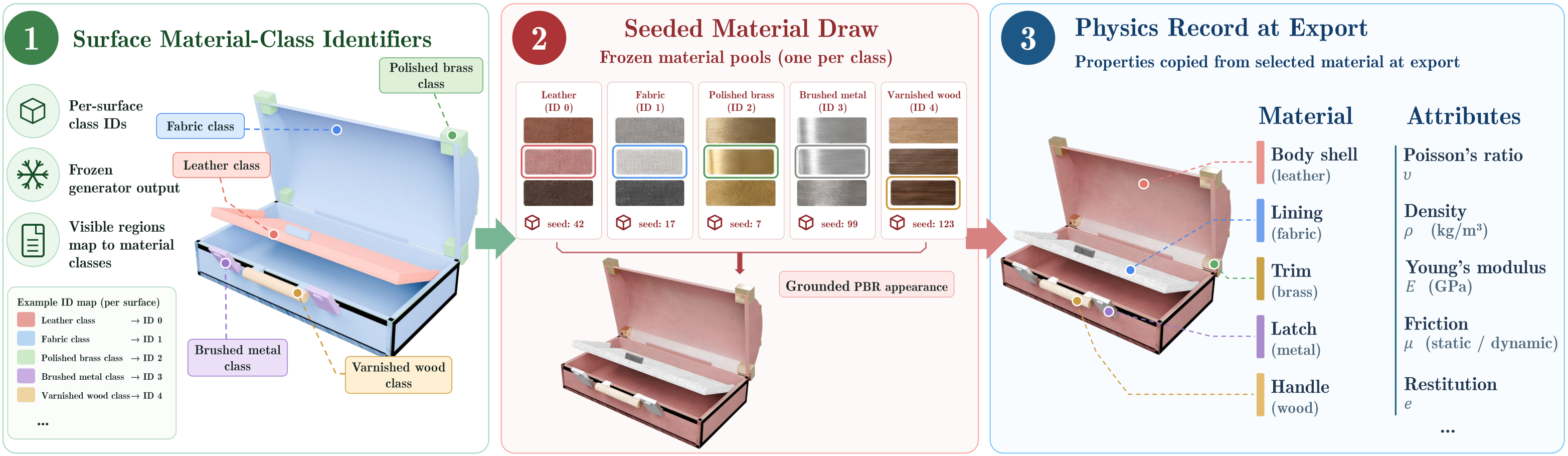}
\caption{How a frozen generator's per-surface material-class identifiers
become concrete materials and physical properties at export: each
identifier selects a frozen material pool, a seeded draw binds one pool
member to the surface, and that material's physical properties are copied
into the asset's physics record.}
\label{fig:material-physics-binding}
\end{figure}

During asset sampling, each surface's class identifier selects a frozen
material pool, and a seeded draw selects an entry from that pool. The selected
appearance is bound to the surface and the corresponding physical properties
are copied into the asset's physics record
(Figure~\ref{fig:material-physics-binding}). This lookup requires no per-asset
model inference and is reproducible from the generator and seed. It also
provides a direct control for material variation: a surface can receive a
different entry from its pool or a different class assignment, while its
appearance and physical annotations are resolved together. 

\begin{figure}[!htbp]
\centering
\includegraphics[width=1.0\textwidth]{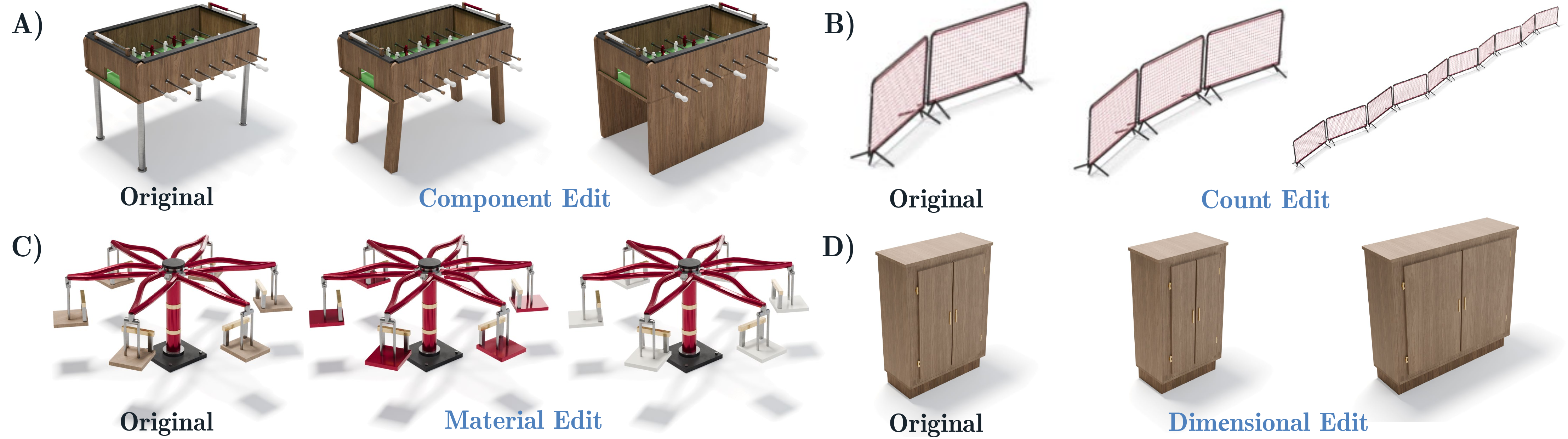}
\caption{Directed 3D editing enabled by agent-generated procedural programs:
(a) structural editing, (b) count editing, (c) material editing, and
(d) dimensional editing.}
\label{fig:directed-3d-editing}
\end{figure}

The generator representation also makes the release editable. An edit changes
the configuration rather than a finished mesh and then re-executes the same
program; no additional per-asset agent call is required.
Figure~\ref{fig:directed-3d-editing} shows four examples: replacing a
component, changing the number of repeated parts, assigning a different
material, and changing a geometric dimension. Dependent placements,
dimensions, and joint parameters are resolved again by the generator, keeping
the edited asset internally consistent. Section~\ref{sec:robotics} uses this
control to adapt retrieved assets to real manipulation targets.

We evaluate the dataset across five aspects:
scale and structural diversity, asset validity,
rest-pose assembly geometry, motion-induced collisions,
and cross-simulator compatibility. Detailed
protocols and results are deferred to Appendices~\ref{app:dataset-evaluation}--
\ref{app:runtime-readiness}.


\providecommand{\name}{\textsc{RFW-A}}
\providecommand{\ours}{\name}
\providecommand{\stackcell}[1]{\shortstack{#1}}
\providecommand{\best}[1]{\ensuremath{\boldsymbol{#1}}}
\providecommand{\tablefont}{\fontsize{8}{9.5}\selectfont}

\section{Experiments}

\subsection{Downstream Articulation Estimation}
\label{sec:downstream-particulate}

\paragraph{Experimental setup.}
We assess whether training with \name-400K improves articulation estimation
from a single 3D mesh. Following Articraft~\citep{articraft2026}, we train
Particulate-B~\citep{particulate2026} on PartNet-Mobility (PM) alone,
PM+Articraft, and PM+\name-400K. All models are trained from random
initialization with identical optimizer settings and a global batch size
of 128, and evaluated at 150K updates. For both augmented datasets,
PM and the additional source are sampled with probabilities 0.1 and 0.9,
respectively.

\paragraph{Evaluation splits and metrics.}
All models share two evaluation splits. The in-domain split contains 255 objects
from PM and Articraft together with additional released objects, covering six
categories seen during training. The category out-of-distribution (OOD) split
contains 774 objects from nine categories excluded from all three training sets.

Following the Particulate evaluation protocol, we report mean IoU (mIoU),
generalized IoU (gIoU), and part-wise Chamfer distance (PC) at rest,
and part-wise gIoU, PC, and whole-object Chamfer distance (OC) at maximum
articulation. Results aggregate three training seeds $\times$ five evaluation
seeds per setting. Training and evaluation details are provided in
Appendix~\ref{app:downstream-particulate}.

\begin{table}[t]
\centering
\caption{Articulation estimation with different training datasets.
Bold marks the best result for each metric within each evaluation split.}
\label{tab:downstream-particulate}
\maintablefont
\setlength{\tabcolsep}{2pt}
\renewcommand{\arraystretch}{1.15}
\newcommand{\relgain}[1]{\textcolor[rgb]{0.00,0.43,0.38}{\,(#1\%)}}
\newcommand{\relchange}[1]{\textcolor[rgb]{0.60,0.22,0.16}{\,(#1\%)}}

\maintablebody{%
\begin{tabular}{@{}>{\raggedright\arraybackslash}m{59pt}%
*{6}{>{\centering\arraybackslash}m{\dimexpr(\textwidth-59pt-12\tabcolsep)/6\relax}}@{}}
\toprule
\multirow{2}{*}{\textbf{Training data}} &
\multicolumn{3}{c}{\textbf{Rest-Pose Segmentation}} &
\multicolumn{3}{c}{\textbf{Articulated Geometry}} \\
\cmidrule(lr){2-4}\cmidrule(lr){5-7}
& \textbf{mIoU}$\uparrow$ & \textbf{gIoU}$\uparrow$ &
\textbf{PC}$\downarrow$ & \textbf{gIoU}$\uparrow$ &
\textbf{PC}$\downarrow$ & \textbf{OC}$\downarrow$ \\
\midrule
\multicolumn{7}{c}{\textbf{In-domain}} \\
\midrule
PM-only &
0.60262 &
0.46364 &
0.09552 &
0.44339 &
0.15760 &
0.01253 \\
PM+Articraft &
0.71442 &
0.62972 &
0.05371 &
0.60164 &
0.10196 &
0.00778 \\
\rowcolor{maintablerow}
PM+\name-400K &
\textbf{0.78873}\relgain{+10.4} &
\textbf{0.73550}\relgain{+16.8} &
\textbf{0.03501}\relgain{+34.8} &
\textbf{0.70295}\relgain{+16.8} &
\textbf{0.06268}\relgain{+38.5} &
\textbf{0.00661}\relgain{+15.0} \\
\midrule
\multicolumn{7}{c}{\textbf{Category OOD}} \\
\midrule
PM-only &
0.35944 &
0.14805 &
0.12802 &
0.13076 &
0.17423 &
0.02481 \\
PM+Articraft &
0.46873 &
0.25616 &
0.11926 &
0.23468 &
0.16176 &
\textbf{0.02343} \\
\rowcolor{maintablerow}
PM+\name-400K &
\textbf{0.47630}\relgain{+1.6} &
\textbf{0.30082}\relgain{+17.4} &
\textbf{0.09660}\relgain{+19.0} &
\textbf{0.27985}\relgain{+19.2} &
\textbf{0.15065}\relgain{+6.9} &
0.02403\relchange{-2.6} \\
\bottomrule
\end{tabular}%
}
\end{table}

\paragraph{Results.}
Table~\ref{tab:downstream-particulate} shows that both generated datasets
improve all six in-domain metrics over PM-only. PM+\name-400K achieves the
best in-domain results on all six metrics. Relative to PM+Articraft, rest-pose
mIoU increases by 7.43 percentage points and articulated gIoU by 10.13 points;
articulated PC falls from 0.10196 to 0.06268 and whole-object OC from 0.00778
to 0.00661. The larger part-level gains indicate that the added assets provide
especially strong supervision for part correspondence and moved-part geometry,
while also improving whole-object reconstruction.

On the category-OOD split, PM+\name-400K again leads on all five part-level
metrics: articulated gIoU improves by 19.2\% and rest-pose PC by 19.0\%
relative to PM+Articraft. The consistent part-level gains across both splits
show that the dataset transfers beyond the categories used for training.

\begin{figure}[t]
\centering
\includegraphics[width=1.0\textwidth]{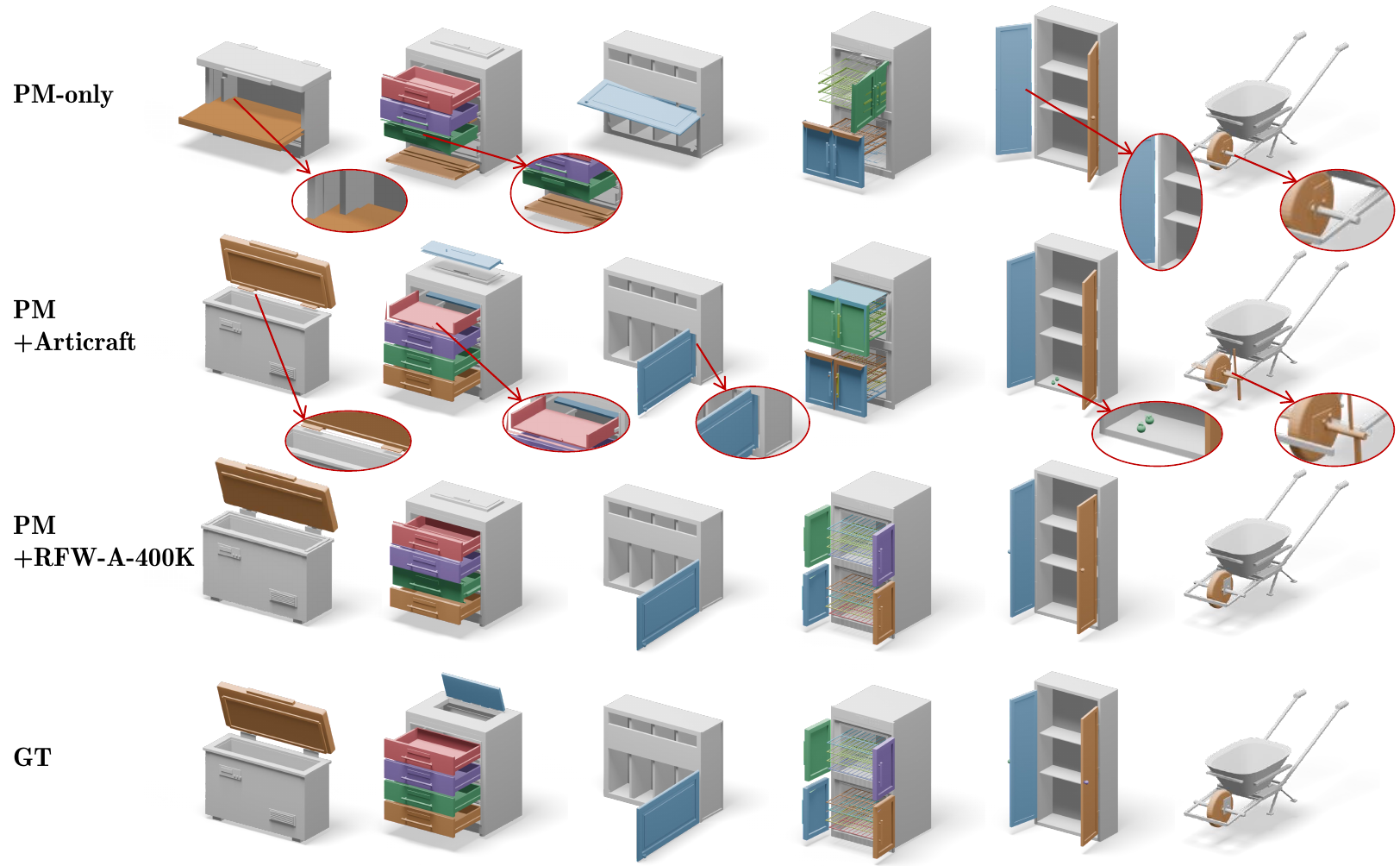}
\caption{Qualitative comparison of articulation estimation. Rows show PM-only,
PM+Articraft, PM+\name-400K, and ground truth (GT). Red arrows and circles
highlight local differences in supports, attachments, and movable parts.}
\label{fig:qualitative-ground-truth}
\end{figure}

Figure~\ref{fig:qualitative-ground-truth} compares articulation estimation results
on six category-matched examples. Each column shows one case,
with PM-only, PM+Articraft, PM+\name-400K, and the ground-truth reference
arranged by row. Red arrows and circles identify local regions where supports,
attachments, or movable parts differ from the reference. The highlighted
examples show that PM-only and PM+Articraft can omit or misplace articulated
geometry, whereas the PM+\name-400K outputs more closely preserve the
reference structure. 

Generator component ablations are reported in Appendix~\ref{app:generator-ablation}:
the full configuration achieves the highest mean Judge score and passes all
strict validity probes.

\section{Robotic Manipulation}
\label{sec:robotics}

We further evaluate whether \name-400K assets can serve as simulation
surrogates for real-world robotic manipulation. For each real target object,
we retrieve the most similar asset from \name-400K and optionally edit its
exposed procedural parameters to better match the target geometry. The
resulting assets are used to train manipulation policies entirely in
simulation, which are then deployed on the corresponding real objects without
real-world training or fine-tuning.

\begin{figure}[t]
\centering
\includegraphics[width=1.0\textwidth]{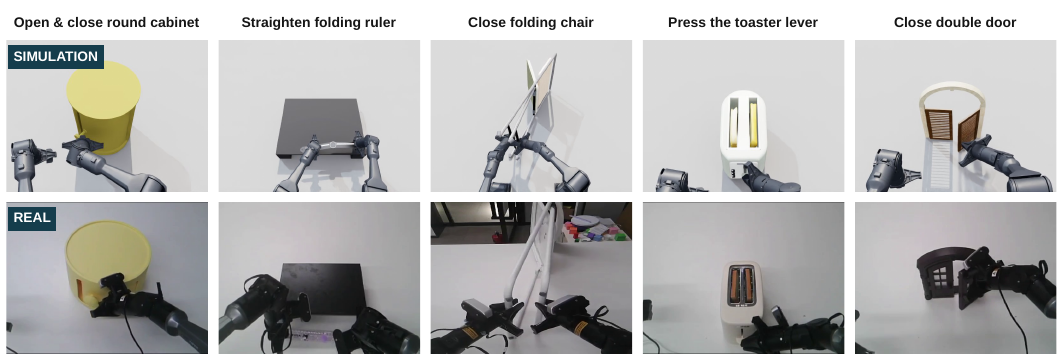}
\caption{\textbf{Sim-to-real robotic manipulation.}
Simulation training environments (top) and corresponding real-world
evaluations (bottom) across five articulated-object manipulation tasks.}
\label{fig:robotics_sim2real}
\end{figure}

\paragraph{Tasks.}
We consider five real-world manipulation tasks with different articulation
structures and interaction patterns:
(1) opening and closing a cabinet door, where the left arm pushes the door
open and the right arm closes it;
(2) straightening a folding ruler;
(3) folding a folding chair to its closed configuration;
(4) pushing down the lever of a toaster; and
(5) closing a pair of double doors.
Together, these tasks cover revolute joints, multi-link articulation, and
object-specific interaction geometries.
Figure~\ref{fig:robotics_sim2real} shows the simulation training environments
and the corresponding real-world evaluations for these five tasks.

\paragraph{Simulation training.}
For each task, we compare three simulation-asset configurations
(Table~\ref{tab:robotics_setup}).
\textbf{Retrieval} uses only the single retrieved \name-400K asset.
\textbf{Retrieval + 4 Edit} augments the retrieved asset with four edited
variants, while \textbf{Retrieval + 9 Edit} augments it with nine edited
variants. The edits modify task-relevant dimensions, part placement, and
articulation parameters through the corresponding procedural generator while
preserving a valid kinematic structure.

Importantly, all three configurations use exactly 500 simulation
demonstrations per task and the same policy architecture and training budget.

\begin{table}[t]
\centering
\caption{\textbf{Simulation configurations for robotic manipulation.}
All settings use exactly 500 simulation demonstrations per task.}
\label{tab:robotics_setup}
\small
\begin{tabular}{lccc}
\toprule
Setting & Retrieved Asset & Edited Variants & Sim. Demos \\
\midrule
Retrieval          & 1 & 0 & 500 \\
Retrieval + 4 Edit & 1 & 4 & 500 \\
Retrieval + 9 Edit & 1 & 9 & 500 \\
\bottomrule
\end{tabular}
\end{table}

\paragraph{Real-world evaluation.}
Each trained policy is evaluated for 10 trials on its corresponding real
object. A trial is counted as successful when the task reaches its predefined
target configuration. We report the number of successful trials for each task
and the average across all five tasks. 

\begin{table}[t]
\centering
\caption{\textbf{Zero-shot sim-to-real manipulation.}
Real-world successes over 10 trials per task. All settings use the same
500 simulation demonstrations per task.}
\label{tab:robotics_results}
\small
\resizebox{\linewidth}{!}{
\begin{tabular}{lcccccc}
\toprule
Setting &
Cabinet &
Folding Chair &
Double Door &
Toaster &
Folding Ruler &
Avg. \\
\midrule
Retrieval          & 5/10 & 7/10 & 8/10 & 7/10 & 4/10 & 6.2/10 \\
Retrieval + 4 Edit & 7/10 & 7/10 & 8/10 & 8/10 & 6/10 & 7.2/10 \\
Retrieval + 9 Edit & 8/10 & 8/10 & 8/10 & 9/10 & 7/10 & 8.0/10 \\
\bottomrule
\end{tabular}}
\end{table}

\paragraph{Results.}
Table~\ref{tab:robotics_results} shows that real-world success improves from
6.2/10 with the retrieved asset alone to 7.2/10 with four edited variants and
8.0/10 with nine edited variants. The gains are largest on the cabinet,
toaster, and folding-ruler tasks, whereas the double-door task remains
unchanged. Since all settings use the same number of simulation demonstrations,
policy architecture, and training budget, these results indicate that
generator-based target adaptation and asset variation improve zero-shot
sim-to-real transfer without requiring additional training data.

\section{Conclusion}
\label{sec:conclusion}

We present RFW-A, an agentic procedural generation framework that constructs reusable category-level generators, enabling efficient asset sampling and editing without per-asset model inference. Using RFW-A, we build RFW-A-400K, comprising over 400K articulated assets across 500+ categories with explicit kinematics, fine-grained materials, and estimated physical properties. Our experiments demonstrate substantially reduced generation costs, high structural and kinematic validity, and reliable execution across four simulators. Beyond asset quality, RFW-A-400K improves articulation estimation on seen and unseen categories and enables zero-shot sim-to-real robotic manipulation through retrieved and procedurally edited assets. These results establish RFW-A as a scalable and controllable approach to articulated asset generation for embodied learning.

\FloatBarrier
\bibliography{iclr2027_conference}
\bibliographystyle{plainnat}

\clearpage
\appendixtableofcontents
\clearpage

\appendix
\section{Appendix}
\label{app:provenance}

\subsection{Generator Construction and Validation}
\label{app:generator-construction}

\begin{algorithm}[H]
    \SetAlFnt{\small}
    \SetAlCapFnt{\normalsize}
    \SetAlCapNameFnt{\normalsize}
    \SetAlgoLined
    \SetAlgoNlRelativeSize{-1}
    \SetCommentSty{texttt}
    \caption{\name generator construction and validation}
    \label{alg:pva}
    \KwIn{category prompt $c$; source assets and programs $\mathcal{A}$;
        repair budget $T$; coverage quota $n$}
    \KwOut{released generator $S_c$ for inference-free sampling
        $x_z=S_c(z)$, or failure}
    $\mathcal{M}_c \leftarrow \operatorname{Extract}(c,\mathcal{A})$
        \tcc*[r]{Component extraction}
    $\Sigma_c^{(0)} \leftarrow \operatorname{Architect}(c,\mathcal{M}_c)$
        \tcc*[r]{Category blueprint}
    $\mathcal{V}_c^{(0)} \leftarrow \operatorname{Choices}(\Sigma_c^{(0)})$
        \tcc*[r]{Declared component choices}
    $S_c^{(0)} \leftarrow \operatorname{Compile}(\Sigma_c^{(0)},\mathcal{M}_c)$
        \tcc*[r]{Resolve parameters, then assemble}
    Probe $\chi_z=\operatorname{Sample}(\Sigma_c^{(0)};z)$ and
    $q_z=\operatorname{Resolve}(\Sigma_c^{(0)},\chi_z)$ over seeds until
    $\#\{z:v\in q_z\}\ge n$ for every $v\in\mathcal{V}_c^{(0)}$\;
    $\mathcal{Z}_{\mathrm{test}}\leftarrow\varnothing$
        \tcc*[r]{Validation cases and retained regression tests}
    \For{$t\leftarrow0$ \KwTo $T$}{
        $\mathcal{Z}_t \leftarrow \mathcal{Z}_{\mathrm{test}}
            \cup\operatorname{Batch}(t)
            \cup\operatorname{Target}(\mathcal{V}_c^{(t)},
            \mathcal{Z}_{\mathrm{test}})$\;
        \ForEach{$z\in\mathcal{Z}_t$}{
            $\chi_z\leftarrow\operatorname{Sample}(\Sigma_c^{(t)};z)$\;
            $q_z\leftarrow\operatorname{Resolve}(\Sigma_c^{(t)},\chi_z)$\;
            $x_z\leftarrow S_c^{(t)}(z)$\;
            $\mathcal{F}_z\leftarrow\operatorname{Check}(x_z)$
                \tcc*[r]{Geometry, URDF, articulation, and collision checks}
        }
        \If{$\left(\forall z\in\mathcal{Z}_t,\ \mathcal{F}_z=\varnothing\right)
            \land\left(\forall v\in\mathcal{V}_c^{(t)},\ \exists z\in\mathcal{Z}_t:
            v\in q_z\right)$}{
            $S_c\leftarrow S_c^{(t)}$\;
            Freeze $S_c$\;
            \Return{$S_c$}\;
        }
        $\mathcal{Z}_{\mathrm{test}}\leftarrow\mathcal{Z}_t$
            \tcc*[r]{Retain failed cases as regression tests}
        \lIf{$t=T$}{\Return{failure}}
        $(\Sigma_c^{(t+1)},S_c^{(t+1)})\leftarrow\operatorname{Repair}
            (\Sigma_c^{(t)},S_c^{(t)},\{\mathcal{F}_z\}_{z\in\mathcal{Z}_t})$
            \tcc*[r]{Blueprint or code repair}
        $\mathcal{V}_c^{(t+1)}\leftarrow\operatorname{Choices}(\Sigma_c^{(t+1)})$
            \tcc*[r]{Recompute coverage after blueprint edits}
    }
    \Return{failure}\;
\end{algorithm}

\FloatBarrier

\subsection{Dataset Versions and Evaluation Sets}
\label{app:evaluation-sets}
Unless noted otherwise, the full-dataset evaluations contain 402{,}410 \name-400K
assets with 1{,}928{,}063 declared non-fixed joints, 2{,}347
PartNet-Mobility assets, 8{,}226 Infinigen-Articulated assets, 4{,}956
SketchMobility assets, 10{,}787 Articraft assets, 2{,}024 PhysX-Mobility
assets, 3{,}544 Artiverse assets, and 3{,}217 LAM assets. The Articraft
evaluation set combines 9{,}996 released packages with 791 GitHub-only source
reconstructions.

The structural-integrity and runtime evaluations use the same Core-200 set.
For each source, we first retain assets with 1--20 declared movable joints and
then sample 200 assets before running either test. Failed assets are not
replaced. The free-standing center-of-mass panel applies an additional
eligibility rule, so its $N$ can be smaller than 200. In that panel,
$N_{\mathrm{phys}}$ is the number of eligible assets with valid
dataset-provided physics.

\subsection{Experimental Setup}
\label{app:dataset-evaluation}

\paragraph{Comparison datasets.}
We compare \name-400K with seven public datasets spanning three construction
paradigms: (a) handcrafted datasets, including
PartNet-Mobility~\citep{sapien2020}, PhysX-Mobility~\citep{physxanything2026},
Artiverse~\citep{artiverse2026}, and
SketchMobility~\citep{yang2026sketch2arti}; (b) procedural datasets,
represented by Infinigen-Articulated~\citep{joshi2025infinigenarticulated};
and (c) agentically generated datasets, including LAM~\citep{lam2026}
and Articraft~\citep{articraft2026}.

\paragraph{Evaluation protocol.}
We evaluate (a) dataset scale and structural diversity
(Appendix~\ref{app:dataset-scale}), (b) asset validity
(Appendix~\ref{app:asset-validity}), (c) rest-pose assembly geometry
(Appendix~\ref{app:assembly-geometry}), (d) collisions during articulation
(Appendix~\ref{app:structural-integrity}), and (e) simulation compatibility
(Appendix~\ref{app:runtime-readiness}). We further
assess the utility of \name-400K as training data for articulation estimation
(Section~\ref{sec:downstream-particulate}). Dataset size and asset validity are
evaluated on the complete releases. For the more expensive geometry and
simulation tests, we randomly sample 200 assets per source with 1--20 declared
movable joints, forming a common evaluation subset that we refer
to as Core-200. Dataset versions and evaluation sets are detailed in Appendix~\ref{app:evaluation-sets}.

\subsection{Dataset Scale and Structural Diversity}
\label{sec:structural-statistics}
\label{app:dataset-scale}

\begin{table}[!htbp]
\centering
\caption{Dataset scale and structural variation.
Joint statistics are computed over successfully parsed assets.
Within-category joint-count variance is averaged equally across each source's
categories. nTED is averaged over all pairs of 250 valid trees per source,
including cross-category pairs. Bold marks the best value in each ranked
column, where a larger value indicates greater scale or structural diversity;
mean and P-90 movable-joint counts are descriptive and are not ranked.}
\label{tab:scale-summary}
\maintablefont
\renewcommand{\arraystretch}{1.08}

\maintablebody{%
\begin{tabular}{lrrrrrrr}
\toprule
\multirow{2}{*}{\textbf{Dataset}} &
\multicolumn{2}{c}{\textbf{Scale}} &
\multicolumn{3}{c}{\textbf{Movable joints}} &
\multicolumn{2}{c}{\textbf{Structural variation}} \\
\cmidrule(lr){2-3}\cmidrule(lr){4-6}\cmidrule(lr){7-8}
& Assets & Parsed links & Mean & P-90 & Multi-joint (\%) & Within-cat. var. & nTED \\
\midrule
PartNet-Mobility      & 2,347   & 16,665    & 5.10 & 9  & 59.86 & 24.52 & 0.411 \\
Infinigen-Articulated & 8,226   & 48,427    & 3.89 & 9  & 73.86 & 2.15  & 0.393 \\
PhysX-Mobility        & 2,024   & 26,016    & 4.89 & 7  & 55.09 & 14.83 & 0.354 \\
Artiverse             & 3,544   & 28,971    & 4.60 & 7  & 77.65 & 12.76 & 0.372 \\
LAM                   & 3,217   & 18,306    & 3.00 & 5  & 53.59 & 2.19  & 0.373 \\
SketchMobility        & 4,956   & 16,733    & 2.22 & 4  & 48.31 & 1.05  & 0.324 \\
Articraft             & 10,787  & 55,107    & 3.71 & 6  & 82.83 & 7.63  & 0.422 \\
\midrule
\rowcolor{maintablerow}
\textbf{RFW-A-400K (Ours)} & \best{402{,}410} & \best{2{,}682{,}098} & 4.79 & 9 &
\best{84.65} & \best{31.02} & \best{0.453} \\
\bottomrule
\end{tabular}%
}
\end{table}

We characterize dataset scale and articulated structure using asset and
parsed-link counts, movable-joint statistics, and kinematic-tree diversity
(Table~\ref{tab:scale-summary}). The mean and 90th percentile (P-90) of
movable-joint counts describe typical complexity and the upper tail.
The multi-joint rate measures the fraction of assets with at least two
movable joints, while within-category joint-count variance captures
variation under each source's category labels.
Following Infinite Mobility~\citep{infinitemobility2025}, we measure
kinematic-tree diversity using normalized tree edit distance (nTED):
the number of edits needed to transform one tree into another, divided
by their combined node count. We average nTED over all pairs of 250 valid
trees per source. This dataset-level measure reflects both category
composition and variation within categories. Counting rules, tree
construction, and sampling details are provided in
Appendix~\ref{app:structural-diversity}.

\name-400K contains 402{,}410 assets, approximately 37 times as many as
Articraft. The scale is accompanied by structural variation: its 84.65\%
multi-joint rate, 31.02 within-category joint-count variance, and 0.453 nTED
are the highest in the comparison, while its P-90 joint count of 9 matches
PartNet-Mobility. Its mean joint count is slightly lower than PartNet-Mobility
(4.79 versus 5.10), so the release expands the number and diversity of
articulated assets without simply shifting toward higher per-asset joint
counts.

\FloatBarrier

\subsection{Asset Validity}
\label{sec:asset-validity}
\label{app:asset-validity}
\label{sec:urdf-validity}
\label{sec:kinematic-executability}

We evaluate resource resolution, non-degenerate motion, and kinematic
validity on each complete release
(Table~\ref{tab:asset-validity-summary}(a)); tree-validity and
joint-specification checks are reported in Appendix~\ref{app:urdf-protocol}.
Each movable joint is tested
at 21 states for valid ranges, finite transforms, non-degenerate motion,
and motion confined to descendant links. An asset passes the kinematic
test only if all movable joints pass; all released assets remain in
the denominator.

\begin{table}[!htbp]
\centering
\caption{Asset evaluation across three test groups. (a) File and kinematic
validity on each complete release. (b) Rest-pose assembly geometry on
Core-200. (c) Import, DoF mapping, inertial completeness, and 1-second
numerical stability on Core-200. Values in (a) and (c) are fractions; (b)
uses percentages. Lower is better for Joint Gap. Bold marks the best value
within each column. See Appendix~\ref{app:collision-protocol} for the
definitions of the geometry metrics and
Appendix Table~\ref{tab:collision-core} for the motion-sweep collision
results.}
\label{tab:asset-evaluation}
\label{tab:asset-validity-summary}
\label{tab:cross-simulator-summary}
\maintablefont
\setlength{\tabcolsep}{2.3pt}
\renewcommand{\arraystretch}{1.08}
\maintablebody{%
\begin{minipage}{\linewidth}
\centering
\begin{tabular}{@{}lccccccccc@{}}
\toprule
\multirow{3}{*}{\textbf{Dataset}} &
\multicolumn{3}{c}{\textbf{(a) Validity}} &
\multicolumn{2}{c}{\textbf{(b) Geometry}} &
\multicolumn{4}{c}{\textbf{(c) Simulation}} \\
& \multicolumn{3}{c}{Full release} &
\multicolumn{2}{c}{Core-200 (\%)} &
\multicolumn{4}{c}{Core-200} \\
\cmidrule(lr){2-4}\cmidrule(lr){5-6}\cmidrule(lr){7-10}
& \shortstack{Resource\\resol.\ $\uparrow$} &
\shortstack{Non-deg.\ $\uparrow$} &
\shortstack{Kinematic\\pass $\uparrow$} &
\shortstack{Joint\\gap P95 $\downarrow$} &
\shortstack{Axis\\support $\uparrow$} &
Import $\uparrow$ &
\shortstack{DoF\\mapping $\uparrow$} &
\shortstack{Stable\\rollout $\uparrow$} &
Inertials $\uparrow$ \\
\midrule
PartNet-Mobility & 0.9859 & 0.9979 & 0.9928 & 39.138 & 57.952 & 0.420 & 0.258 & 0.415 & 0.000 \\
Infinigen-Articulated & 0.9995 & 0.8053 & 0.6944 & 30.223 & 50.591 & 0.930 & 0.903 & 0.930 & 0.000 \\
PhysX-Mobility & \best{1.0000} & 0.2664 & 0.5924 & 45.481 & 47.427 & 0.995 & 0.998 & 0.990 & 0.000 \\
Artiverse & 0.9997 & 0.9704 & 0.9405 & 37.433 & 35.673 & 0.770 & 0.724 & 0.770 & \textbf{1.000} \\
LAM & 0.9714 & 0.8574 & 0.8775 & 54.218 & 32.361 & 0.415 & 0.544 & 0.410 & 0.000 \\
SketchMobility & 0.9903 & 0.8230 & 0.7203 & 22.876 & 68.828 & 0.420 & 0.437 & 0.420 & 0.400 \\
Articraft & 0.9999 & 0.9998 & 0.9999 & 17.716 & 67.755 & 0.920 & 0.946 & 0.915 & 0.710 \\
\midrule
\rowcolor{maintablerow}
\textbf{RFW-A-400K (Ours)} & \best{1.0000} & \best{1.0000} & \best{1.0000} & \best{12.487} & \best{78.828} & \textbf{1.000} & \textbf{0.998} & \textbf{0.995} & \textbf{1.000} \\
\bottomrule
\end{tabular}
\par\smallskip
{\maintablefont\raggedright
Kinematic pass requires every movable joint of an asset to pass; the
non-degeneracy rate is joint-level. Joint gap
measures a robust nearest-surface separation of each parent--child pair and
axis support checks geometry around the declared joint axis; distances are
normalized by the rest-pose bounding-box diagonal. Geometry metrics are
evaluable for 181/200 LAM assets; non-evaluable assets remain failures in
asset-level rates. Simulation import and stable rollout are asset-level and
DoF mapping is joint-level, each requiring a pass in all four simulators;
inertial completeness is evaluated at Genesis import.\par}
\end{minipage}%
}
\end{table}

\name-400K achieves 100\% resource resolution, no degenerate joint motion, and
a 100\% kinematic pass rate on the complete release, compared with 99.99\%
for Articraft. These release-level gates confirm reliable parsing and
kinematic structure at scale; the geometry and motion tests below characterize
how these valid structures behave during articulation.
Detailed checks and failure breakdowns are provided in
Appendices~\ref{app:urdf-protocol} and \ref{app:kinematic-protocol}.

\subsection{Rest-Pose Assembly Geometry}
\label{sec:assembly-geometry}
\label{app:assembly-geometry}

We evaluate rest-pose assembly geometry on Core-200
(Table~\ref{tab:asset-evaluation}(b)) with two metrics: Joint Gap P95, a
robust nearest-surface separation between parent and child parts, and Axis
Support, which checks whether parent and child geometry lies within a fixed
tolerance of the declared joint axis. Distances are normalized by the
rest-pose bounding-box diagonal; Appendix~\ref{app:collision-protocol} gives
the full protocol.

\name-400K has the lowest joint-gap P95 (12.487\%, 5.229 points below
Articraft) and the highest axis support (78.828\%, 11.073 points above
Articraft). The agreement between the two metrics confirms accurate placement
around declared joint interfaces and establishes a strong foundation for the
dynamic collision analysis below.

\subsection{Collisions during Articulation}
\label{sec:structural-integrity}
\label{app:structural-integrity}

We evaluate motion-induced collisions on Core-200
(Appendix Table~\ref{tab:collision-core}) by sweeping each movable
joint through nine uniformly spaced states while keeping other joints
at rest. We report the percentage of sampled poses with no new
collisions and the 95th percentile (P95) of maximum penetration growth,
both measured relative to the rest pose. Penetration growth is normalized
by the rest-pose bounding-box diagonal; local connector contacts are masked.

\name-400K records 84.846\% collision-free poses and a 15.144\%
penetration-growth P95. Together with its strong rest-pose support, these
results demonstrate controlled contact behavior across articulated motion.
The full protocol is given in Appendix~\ref{app:collision-protocol}.

\subsection{Cross-Simulator Compatibility}
\label{sec:runtime-readiness}
\label{app:runtime-readiness}

We test whether the same Core-200 assets can be imported and run in
Genesis~\citep{genesis2024}, PyBullet~\citep{pybullet2021},
MuJoCo~\citep{todorov2012mujoco}, and Isaac Sim~\citep{nvidiaisaacsim}. Table~\ref{tab:cross-simulator-summary}(c)
reports import success, correspondence between declared and simulator
degrees of freedom (DoF), and numerical stability during a 1-second
passive rollout with the root fixed. Each asset or joint must pass the
corresponding check in all four simulators; the reported rates are not
averages across engines. Import and execution failures count as failures.
The table also reports the fraction of assets with complete non-placeholder
inertial parameters at Genesis import.

\name-400K achieves 1.000 import, 0.998 DoF mapping, and 0.995 stable rollout.
Together with Artiverse, it is also one of two datasets with complete inertial
parameters. These rates, together with the fixed-root and free-root checks,
confirm executable structure and physical parameters across simulation
settings. Detailed settings and per-engine results are provided in
Appendix~\ref{app:simulation-protocol}.

\subsection{Material and Physics Annotation Protocol}
\label{app:material-physics}

\paragraph{Taxonomy and library.}
The controlled taxonomy contains 124 leaf material classes grouped into 18
families (for example metal, wood, fabric, and glass). The library contains
8{,}295 PBR materials normalized from BlenderKit, a public PBR material
collection, partitioned into per-class pools; the released snapshot is
read-only and pinned by content hash. Materials whose class assignment
remains flagged for review are excluded from the released pools.

\paragraph{Generator assignment.}
Material-class assignment runs once over the frozen generator corpus,
covering 17{,}640 visual call sites. For each call site, a large
multimodal model proposes candidate classes from the taxonomy, and
deterministic rule checks retain at most two candidates; the assignment
stored on each visual surface of the generator is a single scalar
identifier. Runtime code consumes only this identifier and never interprets
free-text class descriptions.

\paragraph{Physical-property estimation.}
Each library material records six properties in SI units: density
(kg/m\textsuperscript{3}), Young's modulus (Pa), Poisson's ratio, static and
dynamic friction coefficients, and coefficient of restitution. For each
property, a large multimodal model conditioned on the material name, class,
and a rendered preview estimates a value, an uncertainty interval, a
confidence score, and an applicability flag under fixed reference
conditions: density and elastic properties at 23$^\circ$C and 50\% relative
humidity under quasi-static small strain; friction against ground
low-carbon steel (Ra 0.8\,\textmu m) at 10\,kPa and 0.1\,m/s; restitution
from a 25\,mm homogeneous specimen impacting a steel plane at 1\,m/s.
Records must pass hard validity checks (positive density and modulus,
$0\leq\nu<0.5$, $\mu_s\geq\mu_d\geq0$, $0\leq e\leq1$, and
$\mathrm{low}\leq\mathrm{value}\leq\mathrm{high}$) and a risk-tier screen;
records in the most severe tier and entries still flagged for review are
excluded from the release.

These values and uncertainty intervals are model estimates, not physical
measurements. The range checks establish numerical consistency, not accuracy
against measured material properties.

\paragraph{Export-time binding.}
At sampling time, each surface identifier is resolved by a seeded draw from
its class pool, so the same generator version and seed reproduce the same
material and physical values. The exported physics record contains exactly
the six properties above for every bound surface and must align with the
surface keys of the appearance record.

\subsection{Structural Statistics and Diversity}
\label{app:structural-diversity}

\paragraph{Dataset scale and articulation complexity.}
The scale-summary table~\ref{tab:scale-summary} reports these statistics for every
compared source. Its asset count includes every asset in the evaluated
release, while links and movable joints are counted over successfully parsed
records. Movable joints are declared non-fixed XML joints,
including exporter extension joint types. We report the mean
and 90th percentile (P-90) of movable-joint counts. P-90 describes the upper
tail of the distribution without relying on a single maximum. The multi-joint
rate is the fraction of assets with at least two movable joints and describes
how prevalent multi-joint structures are in each dataset.

\paragraph{Within-category joint-count variation.}
Within-category joint variance is the population variance of movable-joint
counts within each category, averaged with equal weight across categories.
It measures variation in articulation complexity within object categories,
which is not captured by the dataset-wide mean. Because the datasets use
different category systems, this statistic is computed under each dataset's
own category labels.

\paragraph{Kinematic tree diversity.}
Different kinematic trees can share the same joint count.
Following the use of tree edit distance in Infinite
Mobility~\citep{infinitemobility2025}, we compare kinematic trees using the
minimum number of node insertions, deletions, and relabelings needed to
transform one tree into another. We contract fixed joints, label each
remaining node by its incoming movable joint type, impose a canonical sibling
order, and use Zhang--Shasha unit costs~\citep{zhang1989simple}.
For trees $T_i$ and $T_j$ with node sets $V_i$ and $V_j$, we compute
\begin{equation}
\operatorname{nTED}(T_i,T_j)
=\frac{\operatorname{TED}(T_i,T_j)}{|V_i|+|V_j|}.
\end{equation}
The normalization reduces the dependence of edit distance on tree size.
We select 250 assets with valid kinematic trees per source by ascending
SHA-256 rank and average nTED over all pairs. Unlike the within-category
comparison in Infinite Mobility, this dataset-wide comparison includes pairs
from different categories. A higher value indicates larger average structural
differences among the sampled trees, reflecting both within-category variation
and category composition.

For \name-400K, the release contains 2{,}682{,}098 parsed links, about 49 times
the 55{,}107 in
Articraft. Its within-category joint-count variance is 31.02, compared with
24.52 for PartNet-Mobility and 7.63 for Articraft. These counts retain each
source's link representation and category definitions.
\FloatBarrier

\subsection{URDF Validity Protocol}
\label{app:urdf-protocol}

Table~\ref{tab:urdf} evaluates each complete release with the same asset-level
rules. Resource Resolution requires every referenced mesh, texture, and
material to exist, be readable, and use a supported format. Valid Tree
requires exactly one root, an acyclic parent--child structure, no orphan
links, and reachability of every link from the root. Joint Specification
requires distinct, existing parent and child links, a finite non-zero axis,
valid joint types, and consistent limits: bounded joints must have finite
limits with lower strictly below upper, whereas continuous joints must not
declare a finite bounded interval.

Collision Coverage requires every declared link, including world or base
links, to contain loadable collision geometry. This all-link rule gives
Infinigen-Articulated zero complete coverage because its dummy world link has no
collision geometry; PhysX-Mobility receives zero because its released URDFs
declare no collision elements. Strict URDF is the per-asset
conjunction of Resource Resolution, Valid Tree, Joint Specification, and
Collision Coverage. Parse failures and incomplete records remain failures in
the complete-release denominator.

\begin{table}[t]
\caption{URDF validity on the complete release of each dataset. Strict URDF is
the per-asset conjunction of resource resolution, tree validity, joint
specification, and collision coverage. All values are fractions; the best
result in each column is bold. See Appendix~\ref{app:urdf-protocol} for the
complete criteria.}
\label{tab:urdf}
\centering
\tablefont
\setlength{\tabcolsep}{3.4pt}
\begin{tabular}{lccccc}
\toprule
\textbf{Dataset} &
\stackcell{\textbf{Resource}\\\textbf{resol.}\ $\uparrow$} &
\stackcell{\textbf{Valid}\\\textbf{tree} $\uparrow$} &
\stackcell{\textbf{Joint spec.}\\\textbf{valid} $\uparrow$} &
\stackcell{\textbf{Collision}\\\textbf{cov.}\ $\uparrow$} &
\stackcell{\textbf{Strict}\\\textbf{URDF} $\uparrow$} \\
\midrule
PartNet-Mobility & 0.9859 & \best{1.0000} & 0.9979 & 0.0000 & 0.0000 \\
Infinigen-Articulated & 0.9995 & \best{1.0000} & \best{1.0000} & 0.0000 & 0.0000 \\
PhysX-Mobility & \best{1.0000} & \best{1.0000} & \best{1.0000} & 0.0000 & 0.0000 \\
Artiverse & 0.9997 & 0.9949 & 0.9989 & 0.9757 & 0.9715 \\
LAM & 0.9714 & 0.9226 & 0.9568 & 0.4597 & 0.4109 \\
SketchMobility & 0.9903 & 0.9986 & 0.9996 & 0.3860 & 0.3858 \\
Articraft & 0.9999 & \best{1.0000} & \best{1.0000} & 0.9988 & 0.9987 \\
\midrule
\textbf{RFW-A-400K (Ours)} & \best{1.0000} & \best{1.0000} & \best{1.0000} & \best{1.0000} & \best{1.0000} \\
\bottomrule
\end{tabular}
\par
\vspace{2pt}
\raggedright\footnotesize PhysX-Mobility declares no collision elements in its
released URDFs. Infinigen-Articulated resolves resources for 8{,}222 of 8{,}226 assets;
its zero complete collision coverage results from the dummy world link.
\end{table}

Under this rule, \name-400K has a Strict URDF rate of 1.0000, followed by
Articraft at 0.9987 and Artiverse at 0.9715. PartNet-Mobility,
Infinigen-Articulated, and PhysX-Mobility have zero complete collision
coverage, despite nearly perfect tree and joint checks. For \name-400K, the
collision-coverage rate is 1.0000 and the joint-specification rate is 1.0000.
\FloatBarrier

\subsection{Kinematic Executability Protocol}
\label{app:kinematic-protocol}

For each declared non-fixed joint, we hold all other joints at their declared
initial states and sample the active joint at $K=21$ uniformly spaced states.
The samples include both limits for bounded joints; continuous joints are
evaluated over $[-\pi,\pi]$. Valid Range requires a finite, non-empty interval
that the evaluator supports. Joint Sweep requires finite forward-kinematic
transforms at every sampled state. Non-degenerate Motion requires the
descendant geometry to exhibit translation or rotation above the
$10^{-6}$ numerical tolerance between sampled states. Subtree Consistency
requires the motion to remain confined to the active joint's descendant
subtree. Joint-level pass is the conjunction of these four tests, and
asset-level pass requires every declared non-fixed joint in an asset to pass.

Valid Range and joint-level pass use all declared non-fixed joints as their
denominator, with unsupported joints and evaluation failures counted as
failures. Joint Sweep and Non-degenerate Motion are computed on joints for
which the corresponding test can be executed; joints excluded from these
conditional columns still fail Valid Range or joint-level pass. Asset-level
pass uses the full asset denominator. Forward-kinematic round-trip error
is retained as a diagnostic but omitted from the headline table because it is
numerically negligible on the measured subset.

\begin{table}[!ht]
\centering
\caption{Kinematic executability on the complete release of each dataset.
Joint-level pass evaluates individual declared movable joints; asset-level
pass requires every movable joint of an asset to pass. All values are
fractions; the best result in each column is bold. See
Appendix~\ref{app:kinematic-protocol} for the sweep and denominator rules.}
\label{tab:kinematic}
\centering
\tablefont
\setlength{\tabcolsep}{4pt}
\begin{tabular}{lccccc}
\toprule
\textbf{Dataset} &
\stackcell{\textbf{Valid range}\\$\uparrow$} &
\stackcell{\textbf{Joint sweep}\\$\uparrow$} &
\stackcell{\textbf{Non-deg.}\\$\uparrow$} &
\stackcell{\textbf{Joint-level}\\\textbf{pass} $\uparrow$} &
\stackcell{\textbf{Asset-level}\\\textbf{pass} $\uparrow$} \\
\midrule
PartNet-Mobility & 0.9960 & \best{1.0000} & 0.9979 & 0.9939 & 0.9928 \\
Infinigen-Articulated & \best{1.0000} & \best{1.0000} & 0.8053 & 0.8053 & 0.6944 \\
PhysX-Mobility & \best{1.0000} & \best{1.0000} & 0.2664 & 0.2664 & 0.5924 \\
Artiverse & 0.9993 & 0.9939 & 0.9704 & 0.9697 & 0.9405 \\
LAM & 0.9949 & 0.8604 & 0.8574 & 0.8530 & 0.8775 \\
SketchMobility & 0.9998 & 0.9991 & 0.8230 & 0.8229 & 0.7203 \\
Articraft & \best{1.0000} & 0.9998 & 0.9998 & 0.9998 & 0.9999 \\
\midrule
\textbf{RFW-A-400K (Ours)} & \best{1.0000} & \best{1.0000} & \best{1.0000} & \best{1.0000} & \best{1.0000} \\
\bottomrule
\end{tabular}
\par
\vspace{2pt}
\raggedright
\footnotesize
Sweep and non-degenerate-motion rates use evaluable joints. Unsupported joints
and evaluation failures remain failures in Valid Range and joint-level pass.
\end{table}

The joint-level and asset-level pass rates for \name-400K are 1.0000 and
1.0000, respectively, compared with 0.9998 and 0.9999 for Articraft.
The largest reduction for Infinigen-Articulated, PhysX-Mobility, and
SketchMobility occurs in the non-degenerate-motion check; LAM also has a
lower sweep pass rate.
\FloatBarrier

\subsection{Geometric Support and Collision Protocol}
\label{app:collision-protocol}

Table~\ref{tab:collision-core} evaluates Core-200 by computing each metric per
asset and then averaging across assets. At the rest pose, Joint Support samples
the surfaces of the
parent and child near the joint origin and requires support in both
directions. The reported value is the mean, across assets, of the fraction of
evaluable joints that pass. All-joint Support is a stricter asset-level metric
that
requires a single root and requires every joint to be evaluable and supported.
Joint Gap computes a robust nearest-surface separation for each parent--child
pair; Joint Gap P95 reports the 95th percentile over the asset summaries.
Axis Support tests whether geometry from both the parent and child lies within
the fixed tolerance of a bounded segment along the declared joint axis.

For the motion tests, each movable joint is swept through nine uniformly
spaced states while all other joints remain at the rest state. Pose Axis
Support is the fraction of those states retaining support around the declared
axis, averaged per asset. Collision-Free Pose Rate is the fraction of sampled
states without motion-induced collision. To avoid treating existing
rest-pose intersections as new failures, this test measures penetration growth
relative to the rest state and masks local contacts at the active joint that
serve as connector contacts. Penetration Growth P95 reports the 95th
percentile of the maximum additional penetration over joints and poses.
Distance quantities are divided by the rest-pose bounding-box diagonal.
Non-evaluable assets are counted as failures in asset-level metrics; the numbers
of evaluable assets for exceptional sources are reported below the table.

\begin{table}[H]
\caption{Geometric support and collision behavior on Core-200. Rest-pose
metrics evaluate assembly support, part separation, and axis placement;
motion metrics evaluate support retention and additional penetration over
nine joint states. All values are percentages. Lower is better for Joint Gap
and Penetration Growth; the best result in each column is bold. See
Appendix~\ref{app:collision-protocol} for definitions.}
\label{tab:collision-core}
\centering
\tablefont
\setlength{\tabcolsep}{3pt}
\begin{tabular}{lccccccc}
\toprule
\multirow{2}{*}{\textbf{Dataset}} &
\multicolumn{4}{c}{\textbf{Rest-Pose Geometry}} &
\multicolumn{3}{c}{\textbf{Motion Sweep ($K=9$)}} \\
\cmidrule(lr){2-5}\cmidrule(lr){6-8}
& \stackcell{\textbf{Joint}\\\textbf{Support} $\uparrow$} &
\stackcell{\textbf{All-joint}\\\textbf{Support} $\uparrow$} &
\stackcell{\textbf{Joint Gap}\\\textbf{P95} $\downarrow$} &
\stackcell{\textbf{Axis}\\\textbf{Support} $\uparrow$} &
\stackcell{\textbf{Pose Axis}\\\textbf{Support} $\uparrow$} &
\stackcell{\textbf{Collision-free}\\\textbf{Poses} $\uparrow$} &
\stackcell{\textbf{Penetr.\ Growth}\\\textbf{P95} $\downarrow$} \\
\midrule
PartNet-Mobility & 20.124 & 15.500 & 39.138 & 57.952 & 48.688 & 83.023 & 29.402 \\
Infinigen-Articulated & \best{60.344} & \best{42.500} & 30.223 & 50.591 & 39.028 & 76.840 & 37.200 \\
PhysX-Mobility & 13.148 & 9.000 & 45.481 & 47.427 & 39.754 & 78.152 & 36.497 \\
Artiverse & 16.657 & 5.500 & 37.433 & 35.673 & 17.611 & 84.599 & 32.763 \\
LAM & 21.933 & 10.500 & 54.218 & 32.361 & 17.623 & 71.075 & 43.443 \\
SketchMobility & 28.514 & 22.500 & 22.876 & 68.828 & \best{65.389} & 86.711 & \best{11.155} \\
Articraft & 42.698 & 16.000 & 17.716 & 67.755 & 43.965 & \best{87.253} & 17.730 \\
\midrule
\textbf{RFW-A-400K (Ours)} & 51.424 & 21.500 & \best{12.487} & \best{78.828} & 53.954 & 84.846 & 15.144 \\
\bottomrule
\end{tabular}
\par
\vspace{2pt}
\raggedright\footnotesize A small number of assets are non-evaluable for individual
metrics (LAM: 181/200 for support, gap, and axis, 180/200 for pose-based
metrics; PartNet-Mobility and PhysX-Mobility: 199/200 for pose-based
metrics); they are counted as failures in the asset-level rates.
\end{table}

\name-400K has the lowest Joint Gap P95 (12.487\% of the object diagonal)
and the highest rest-pose Axis Support (78.828\%) under these diagnostics.
Its Joint Support and Pose Axis Support rates are 51.424\% and 53.954\%.
Infinigen-Articulated has the highest Joint Support and All-joint Support,
whereas SketchMobility has the highest Pose Axis Support. All-joint Support
ranges from 5.5\% to 42.5\% across sources. These local support tests depend
on geometric thresholds and the placement of joint frames; semantic correctness
requires evidence beyond this test.
\FloatBarrier

\subsection{Simulation Compatibility and Stability Protocol}
\label{app:simulation-protocol}

Table~\ref{tab:table5_simulation_readiness} separates fixed-root execution
in Genesis, free-root stability on the eligible subset, and the
cross-simulator intersection reported in the main paper.

\paragraph{Genesis import and rollout (panel a1).}
Each asset is reset three times to its canonical URDF neutral state. Bounded
coordinates are clamped to their valid ranges during reset. The asset is then
simulated in Genesis for 10 seconds at 240 Hz with gravity, contacts,
self-collision, the physical parameters in each asset manifest, a fixed root,
and zero applied joint force. Import and DoF mapping require finite simulator
states and consistent
canonical mappings. Genesis simulation validity additionally requires finite
states and observed poses at all 2,400 steps in every reset repetition.
Trajectory coverage is the fraction of evaluated articulation trajectories
observed for the complete rollout.

Complete non-placeholder inertials require every required non-root dynamic link
to have a positive finite mass, a finite center of mass, and a valid
positive-definite inertia satisfying the rigid-body triangle inequality. An
asset is not counted as complete when all required links use the exact unit
placeholder inertia convention.

\paragraph{Free-root CoM stability (panel a2).}
For this panel, the root is free and all movable joints are held at
canonical $q=0$. The asset is positioned with the collision bounding-box
bottom aligned to an infinite plane, with a 2 mm initial clearance. Gravity is
$[0,0,-9.81]\,\mathrm{m/s^2}$, the timestep is $1/240$ s, and the simulation
runs for 10 seconds. During the first 0.5 seconds, ground-contact positions
within the contact tolerance are deduplicated and used to construct a convex
support polygon in the ground plane.

Mass/CoM Coverage is $N_{\mathrm{phys}}/N$, where valid physical parameters
require finite mass and center-of-mass values. CoM Support Coverage is the fraction of
$N_{\mathrm{phys}}$ assets with a non-degenerate contact support polygon. CoM
Static Stability is also computed with $N_{\mathrm{phys}}$ as its denominator;
it requires a non-negative margin (within the prescribed tolerance), an
available support polygon, and a passing full 10-second rollout. If $N_{\mathrm{phys}} = 0$, the conditional CoM metrics
are reported as ``N/A'', and these cases are not counted
as physical-instability failures. The signed CoM support margin is
the distance from the horizontal center-of-mass projection to the
support-polygon boundary, positive inside the polygon and negative outside;
the normalized form divides by the support-polygon diameter. The rollout
criterion limits transient tilt to $60^\circ$, final-window tilt to
$15^\circ$, and root drop to one object bounding-box height.

The CoM panel is a diagnostic physical-stability evaluation. To support
heterogeneous asset formats, the runtime uses visual geometry when collision
geometry is unavailable, repairs degenerate collision meshes when needed, and
recomputes inertias for simulator execution. Mass/CoM Coverage uses only
parameters provided by the source dataset, while the two conditional metrics
exclude assets without valid source mass and CoM values. Mass and CoM values
filled in by Genesis are retained only in the supplementary diagnostic output.
This separation distinguishes physical-parameter coverage from simulator
execution.

\paragraph{Cross-simulator evaluation (panel b).}
The same canonical asset selection and joint mapping are evaluated in Genesis,
PyBullet, MuJoCo, and Isaac Sim. For each asset or joint, a pass is recorded only when the
corresponding criterion passes in all four simulators. The reported import, DoF-mapping, and stable-rollout rates
are computed from the intersection of passing cases
across all four simulators. The
stable-rollout criterion is a 1-second (240-step at 240 Hz) zero-force passive
fixed-root rollout requiring finite states and observed poses throughout.
Table~\ref{tab:cross_sim_per_engine} reports the per-engine rates behind the
intersection.

\paragraph{Failure and reporting policy.}
Worker errors, incomplete rollouts, and missing mappings are counted as
failures. Unavailable contact support and missing source physical parameters
are reflected in their corresponding coverage metrics. Detailed per-asset
records, protocol hashes, collision-repair logs, and failure categories are
retained with the released experiment artifacts.

\begin{table}[H]
  \centering
  \caption{
    Runtime compatibility and physical stability on Core-200.
    Panel (a1) evaluates import and rollout behavior in Genesis;
    panel (a2) evaluates free-root center-of-mass stability in Genesis;
    and panel (b) requires each asset or joint to pass the corresponding test
    in Genesis, PyBullet, MuJoCo, and Isaac Sim. All reported rates are fractions.
  }
  \label{tab:table5_simulation_readiness}

  \vspace{2pt}
  \textbf{(a1) Genesis import and rollout}

  \begingroup
  \scriptsize
  \setlength{\tabcolsep}{3.2pt}
  \renewcommand{\arraystretch}{1.12}
  \resizebox{\textwidth}{!}{%
    \begin{tabular}{@{}lrrrrrr@{}}
      \toprule
      Dataset &
      $N$ &
      \shortstack{Import $\uparrow$} &
      \shortstack{DoF mapping $\uparrow$} &
      \shortstack{Complete non-placeholder\\inertials $\uparrow$} &
      \shortstack{Genesis 10\,s simulation\\validity $\uparrow$} &
      \shortstack{Trajectory coverage $\uparrow$} \\
      \midrule
      PartNet-Mobility & 200 & 0.950 & 0.920 & 0.000 & 0.950 & 0.000 \\
      Infinigen-Articulated & 200 & 0.930 & 0.903 & 0.000 & 0.930 & 0.000 \\
      PhysX-Mobility & 200 & \textbf{1.000} & \textbf{1.000} & 0.000 &
      0.990 & \textbf{1.000} \\
      Artiverse & 200 & 0.995 & 0.992 & \textbf{1.000} & 0.960 & 0.955 \\
      LAM & 200 & 0.875 & 0.989 & 0.000 & 0.870 & 0.958 \\
      SketchMobility & 200 & 0.995 & 0.990 & 0.400 & \textbf{0.995} & 0.801 \\
      Articraft & 200 & 0.980 & 0.995 & 0.710 & 0.975 & \textbf{1.000} \\
      \textbf{RFW-A-400K (Ours)} & 200 & \textbf{1.000} & \textbf{1.000} &
      \textbf{1.000} & \textbf{0.995} & 0.994 \\
      \bottomrule
    \end{tabular}%
  }
  \endgroup

  \vspace{5pt}
  \textbf{(a2) Genesis free-root center-of-mass stability}

  \begingroup
  \small
  \setlength{\tabcolsep}{3.8pt}
  \renewcommand{\arraystretch}{1.08}
  \begin{tabular}{@{}lrrrrr@{}}
    \toprule
    Dataset & Eligible $N$ & $N_{\mathrm{phys}}$ &
    \shortstack{Mass/CoM cov.\ $\uparrow$} &
    \shortstack{CoM support cov.$^{\dagger}$\ $\uparrow$} &
    \shortstack{CoM static stability$^{\dagger}$\ $\uparrow$} \\
    \midrule
    PartNet-Mobility     & 158 & 0   & 0.000         & N/A            & N/A \\
    Infinigen-Articulated        & 170 & 0   & 0.000         & N/A            & N/A \\
    PhysX-Mobility       & 130 & 130 & 1.000         & 0.992         & 0.846 \\
    Artiverse            & 160 & 160 & 1.000         & 0.994         & 0.919 \\
    LAM                  & 75  & 4   & 0.053         & 1.000         & 0.750 \\
    SketchMobility       & 163 & 67  & 0.411         & 0.940         & 0.836 \\
    Articraft            & 159 & 114 & 0.717         & 0.974         & 0.886 \\
    \textbf{RFW-A-400K (Ours)}  & 156 & 156 & 1.000         & 0.994         & 0.891 \\
    \bottomrule
  \end{tabular}
  \endgroup

  \vspace{5pt}
  \textbf{(b) Cross-simulator evaluation}

  \begingroup
  \small
  \setlength{\tabcolsep}{9pt}
  \renewcommand{\arraystretch}{1.08}
  \begin{tabular}{@{}lrrr@{}}
    \toprule
    Dataset &
    \shortstack{Import $\uparrow$} &
    \shortstack{DoF mapping $\uparrow$} &
    \shortstack{Stable rollout $\uparrow$} \\
    \midrule
    PartNet-Mobility     & 0.420         & 0.258         & 0.415 \\
    Infinigen-Articulated        & 0.930         & 0.903         & 0.930 \\
    PhysX-Mobility       & 0.995         & \textbf{0.998}         & 0.990 \\
    Artiverse            & 0.770         & 0.724         & 0.770 \\
    LAM                  & 0.415         & 0.544         & 0.410 \\
    SketchMobility       & 0.420         & 0.437         & 0.420 \\
    Articraft            & 0.920         & 0.946         & 0.915 \\
    \textbf{RFW-A-400K (Ours)}  & \textbf{1.000} & \textbf{0.998} & \textbf{0.995} \\
    \bottomrule
  \end{tabular}
  \endgroup

  \vspace{3pt}
  \begin{minipage}{0.98\textwidth}
    \footnotesize
    $^{\dagger}$Metrics in (a2) are conditional on $N_{\mathrm{phys}}$;
    best values are not highlighted because the conditional sample sizes
    differ. Each value in (b) requires passing the corresponding test in all
    four simulators. Appendix~\ref{app:simulation-protocol} gives the complete
    protocol.
  \end{minipage}
\end{table}

\begin{table}[H]
  \centering
  \caption{
    Per-engine cross-simulator rates on Core-200: native import success,
    canonical DoF mapping (joint-level), and passive stable rollout
    (asset-level), each evaluated independently per engine. The main-paper
    panel (b) reports the per-asset and per-joint intersection across all
    four engines, not an average of these per-engine rates.
  }
  \label{tab:cross_sim_per_engine}
  \begingroup
  \scriptsize
  \setlength{\tabcolsep}{3.2pt}
  \renewcommand{\arraystretch}{1.12}
  \resizebox{\textwidth}{!}{%
    \begin{tabular}{@{}lrrrrrrrrrrrr@{}}
      \toprule
      & \multicolumn{3}{c}{Genesis} & \multicolumn{3}{c}{PyBullet} & \multicolumn{3}{c}{MuJoCo} & \multicolumn{3}{c}{Isaac Sim} \\
      \cmidrule(lr){2-4}\cmidrule(lr){5-7}\cmidrule(lr){8-10}\cmidrule(lr){11-13}
      Dataset & Imp.\ $\uparrow$ & DoF $\uparrow$ & Stable $\uparrow$ & Imp.\ $\uparrow$ & DoF $\uparrow$ & Stable $\uparrow$ & Imp.\ $\uparrow$ & DoF $\uparrow$ & Stable $\uparrow$ & Imp.\ $\uparrow$ & DoF $\uparrow$ & Stable $\uparrow$ \\
      \midrule
      PartNet-Mobility & 0.950 & 0.920 & 0.945 & 0.985 & 0.994 & 0.985 & 0.420 & 0.258 & 0.420 & 0.975 & 0.984 & 0.970 \\
      Infinigen-Articulated & 0.930 & 0.903 & 0.930 & 1.000 & 1.000 & 1.000 & 0.995 & 0.995 & 0.995 & 1.000 & 1.000 & 1.000 \\
      PhysX-Mobility & 1.000 & 1.000 & 0.995 & 1.000 & 1.000 & 0.995 & 1.000 & 1.000 & 0.995 & 0.995 & 0.998 & 0.990 \\
      Artiverse & 0.995 & 0.992 & 0.960 & 1.000 & 1.000 & 1.000 & 0.850 & 0.855 & 0.845 & 0.915 & 0.839 & 0.915 \\
      LAM & 0.875 & 0.989 & 0.870 & 0.880 & 0.994 & 0.880 & 0.440 & 0.546 & 0.420 & 0.920 & 1.000 & 0.895 \\
      SketchMobility & 0.995 & 0.990 & 0.995 & 0.550 & 0.680 & 0.550 & 0.855 & 0.749 & 0.855 & 1.000 & 1.000 & 1.000 \\
      Articraft & 0.980 & 0.995 & 0.975 & 0.980 & 1.000 & 0.980 & 0.920 & 0.951 & 0.920 & 0.980 & 1.000 & 0.980 \\
      \textbf{RFW-A-400K (Ours)} & 1.000 & 1.000 & 1.000 & 1.000 & 1.000 & 1.000 & 1.000 & 0.998 & 0.995 & 1.000 & 1.000 & 1.000 \\
      \bottomrule
    \end{tabular}%
  }
  \endgroup
\end{table}

All Core-200 assets from \name-400K import into Genesis, map their DoFs, and
pass the complete-inertials check. Their simulation-validity and
trajectory-coverage rates are 0.995 and 0.994. In the free-root panel, all
156 eligible \name-400K assets have source-provided mass and CoM values;
conditional support coverage is 0.994 and static stability is 0.891,
compared with 0.919 static stability for Artiverse.
\FloatBarrier

\subsection{All-Available Feature-Space Analysis}
\label{app:feature-consistency}
\label{sec:visual-diversity}

Beyond structural validation, we examine whether images with the same category
label cluster in feature space under standardized rendering. We render five
datasets in Blender using the same image resolution, camera direction and field
of view, lighting, background, color management, and Cycles settings. Camera
distance is adjusted only according to the object bounding sphere so that each
object remains fully visible. We retain all successfully rendered images
in the evaluation subset, including those from
categories with only one render.
No duplicate images are added. This all-available cohort
contains 1{,}565 rendered category labels and 5{,}725 renders: 531 categories
(2{,}655 renders) for \name-400K, 244 (1{,}193) for Articraft, and 660 rendered
LAM labels (1{,}279 renders); Table~\ref{tab:table0} reports 650 LAM release
categories under the canonical taxonomy. Artiverse contributes 84 (368), and
PartNet-Mobility 46 (230).

Table~\ref{tab:all-available-feature-space} reports the Top-1 same-class
nearest-neighbor rate, the Top-4 same-class fraction among the four nearest
neighbors, and the cosine silhouette. Images from singleton categories are
excluded only from metrics that require a positive same-class pair; they remain
in the t-SNE figures. The resulting metric-image counts are 2{,}655 for
\name-400K, 363 for Artiverse, 230 for PartNet-Mobility, 1{,}187 for Articraft,
and 860 for LAM.

\name-400K obtains the highest DINOv2 Top-1 rate (0.881), Top-4 rate (0.704), and
cosine silhouette (0.289), as well as the highest CLIP Top-1 rate (0.592).
Artiverse has the higher CLIP Top-4 rate (0.395) and cosine silhouette (0.074,
versus 0.064 for \name-400K).

\begin{figure}[!htbp]
\centering
\begin{minipage}{1.0\textwidth}
\begin{minipage}[t]{0.46\textwidth}
\centering
\includegraphics[width=\linewidth]{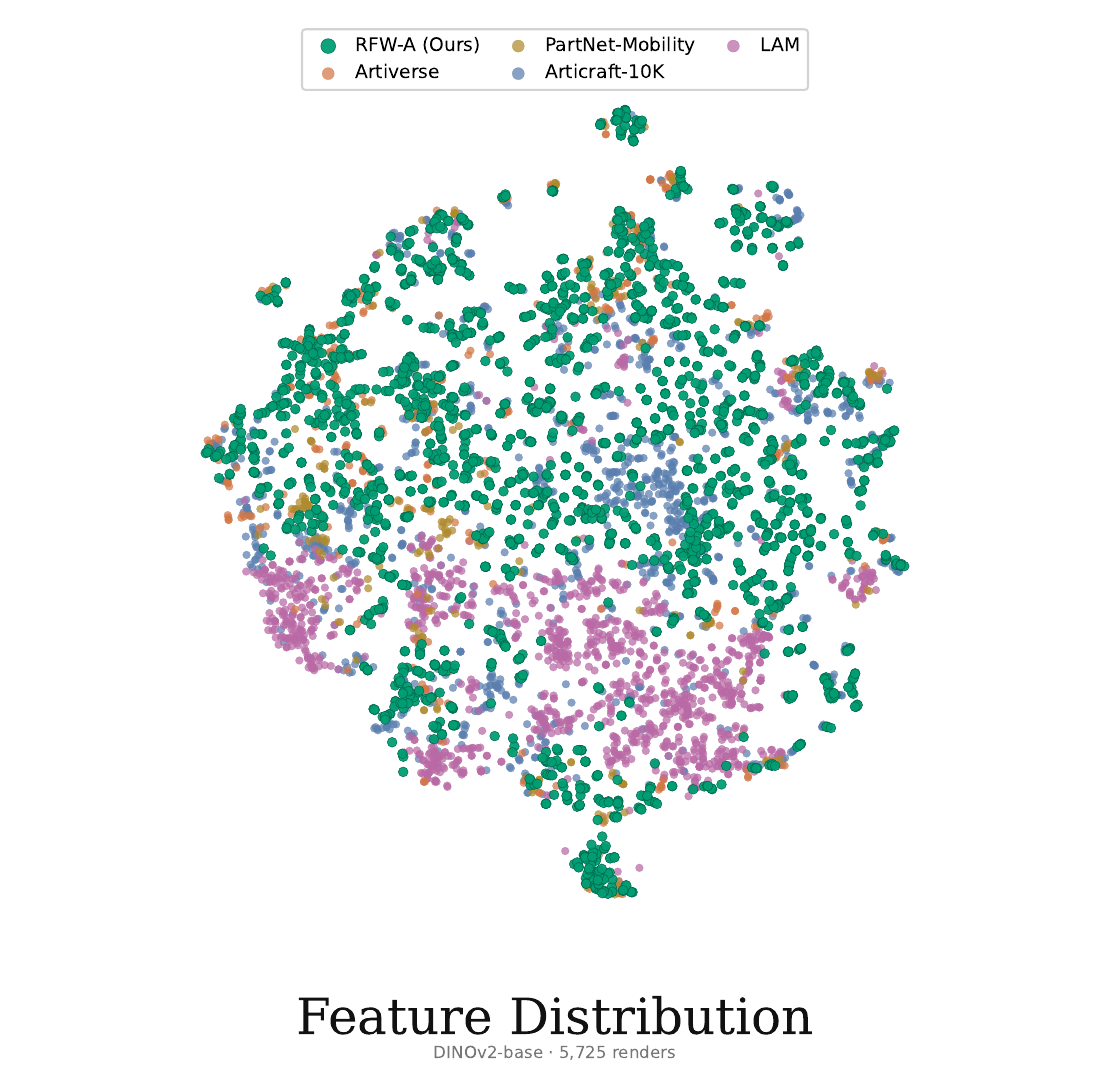}
\end{minipage}\hfill
\begin{minipage}[t]{0.46\textwidth}
\centering
\includegraphics[width=\linewidth]{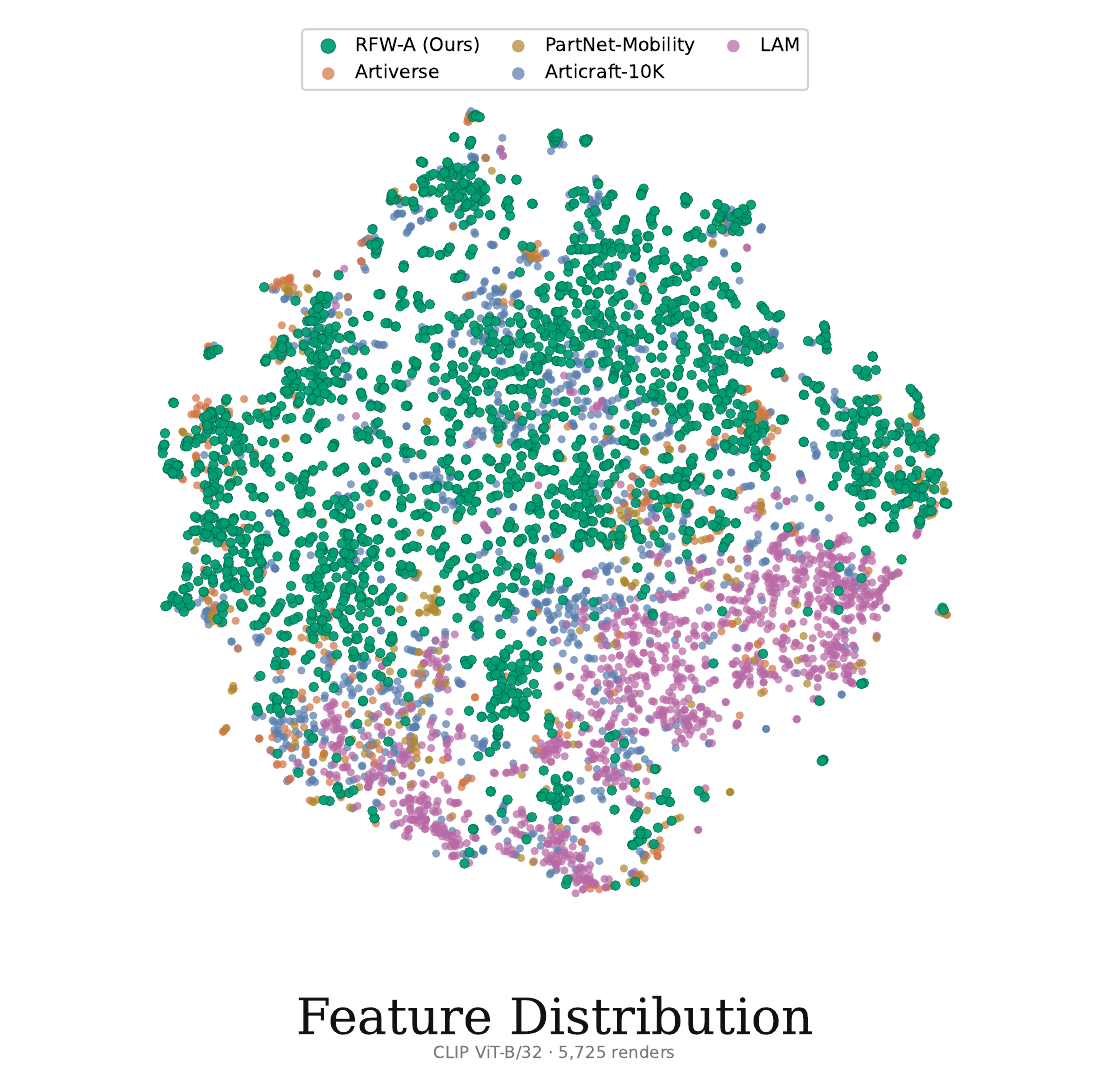}
\end{minipage}
\end{minipage}
\caption{Joint all-available t-SNE overviews in DINOv2-base (left) and CLIP
ViT-B/32 (right) feature spaces. Within each encoder, panels use that
encoder's joint embedding coordinates and axis limits; the two encoders are
embedded separately. Colors identify datasets (\name{} uses larger outlined markers); all renders,
including classes with fewer than five renders, are shown.}
\label{fig:all-tsne-joint}
\end{figure}

\begin{figure}[!htbp]
\centering
\includegraphics[width=0.98\textwidth]{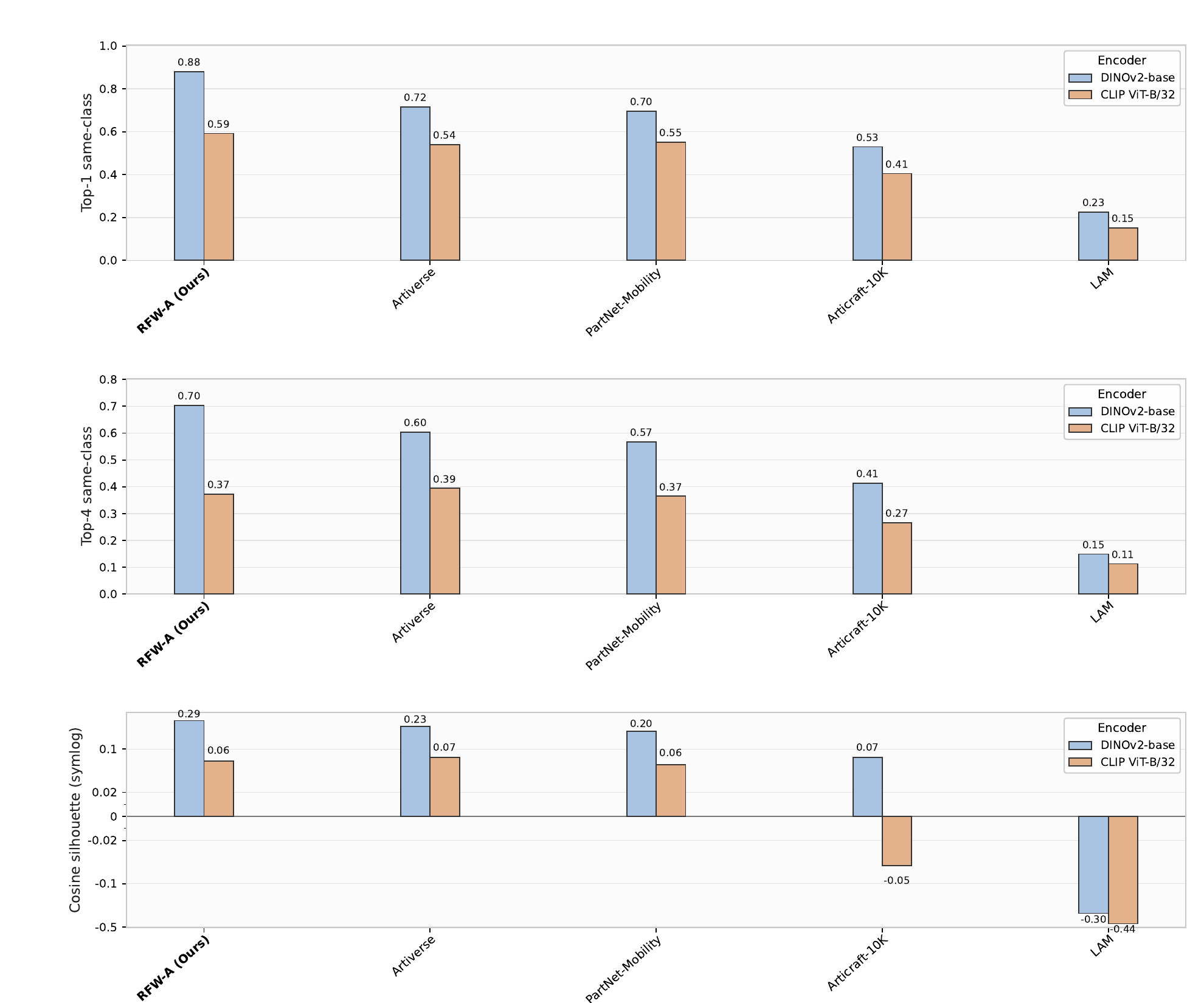}
\caption{All-available feature-space metrics reported in
Table~\ref{tab:all-available-feature-space}. Singleton classes are excluded
only from metrics requiring a positive same-class pair; triangles mark the best
value within each encoder.}
\label{fig:all-metrics}
\end{figure}

Figure~\ref{fig:all-tsne-joint} provides side-by-side joint t-SNE overviews for
DINOv2 and CLIP. Within each encoder, all panels reuse that encoder's joint
embedding coordinates and axis limits; the DINOv2 and CLIP coordinates are
fitted separately. Figure~\ref{fig:all-metrics} summarizes the three
all-available metrics, and
Figures~\ref{fig:all-tsne-pva}--\ref{fig:all-tsne-lam} repeat the side-by-side
comparison for each dataset.

\begin{table}[t]
\caption{Feature-space class consistency under standardized Blender rendering
using all successfully rendered assets. \textbf{Classes} and \textbf{renders}
include every available class and image; metric $N$ excludes only images from
singleton classes when a positive same-class pair is required. Top-1 and Top-4
are the same-class fractions among the nearest one and four neighbors,
respectively. The cosine silhouette is $(b_i-a_i)/\max(a_i,b_i)$ over
L2-normalized features. The best value in each metric column is bold.}
\label{tab:all-available-feature-space}
\centering
\footnotesize
\setlength{\tabcolsep}{3.2pt}
\renewcommand{\arraystretch}{1.08}
\resizebox{\textwidth}{!}{%
\begin{tabular}{lccccccccc}
\toprule
\textbf{Dataset} &
\stackcell{\textbf{Classes}} &
\stackcell{\textbf{Renders}} &
\stackcell{\textbf{Metric}\\\textbf{$N$}} &
\stackcell{\textbf{DINOv2}\\\textbf{Top-1} $\uparrow$} &
\stackcell{\textbf{DINOv2}\\\textbf{Top-4} $\uparrow$} &
\stackcell{\textbf{DINOv2}\\\textbf{silh.} $\uparrow$} &
\stackcell{\textbf{CLIP}\\\textbf{Top-1} $\uparrow$} &
\stackcell{\textbf{CLIP}\\\textbf{Top-4} $\uparrow$} &
\stackcell{\textbf{CLIP}\\\textbf{silh.} $\uparrow$} \\
\midrule
PartNet-Mobility & 46 & 230 & 230 & 0.696 & 0.567 & 0.196 & 0.552 & 0.365 & 0.056 \\
Artiverse & 84 & 368 & 363 & 0.716 & 0.604 & 0.233 & 0.540 & \best{0.395} & \best{0.074} \\
LAM & 660 & 1{,}279 & 860 & 0.226 & 0.149 & -0.301 & 0.152 & 0.113 & -0.442 \\
Articraft & 244 & 1{,}193 & 1{,}187 & 0.530 & 0.413 & 0.073 & 0.405 & 0.266 & -0.051 \\
\midrule
\textbf{RFW-A-400K (Ours)} & \textbf{531} & \textbf{2{,}655} & \textbf{2{,}655} &
\best{0.881} & \best{0.704} & \best{0.289} & \best{0.592} & 0.373 & 0.064 \\
\bottomrule
\end{tabular}%
}
\end{table}

\begingroup
\captionsetup{hypcap=false}
\begin{figure}[!htbp]
\input{figures/appendix/fig10_rfwa_tsne}
\par\vspace{8pt}
\input{figures/appendix/fig11_artiverse_tsne}
\par\vspace{8pt}
\input{figures/appendix/fig12_articraft_tsne}
\end{figure}
\begin{figure}[!htbp]
\input{figures/appendix/fig13_partnet_mobility_tsne}
\par\vspace{8pt}
\input{figures/appendix/fig14_lam_tsne}
\end{figure}
\endgroup

\subsection{Downstream Articulation Estimation}
\label{app:downstream-particulate}

\paragraph{Training configuration.}
All three models use the public Particulate-B architecture and the
same loss, augmentation, and PartField features. We train each setting from
random initialization with three independent training seeds for 180{,}000
optimizer updates on eight GPUs
in FP32, with a global batch size of 128. AdamW uses
$\beta=(0.9,0.999)$, weight decay 0.01, and $\epsilon=10^{-8}$.
The learning rate reaches $6.4\times10^{-5}$ after 1{,}000 warm-up
updates and follows a cosine schedule to $6.4\times10^{-6}$.
We report the fixed 150K checkpoints for all three models and evaluation
splits.

The training sets contain 2{,}087 PM assets, 9{,}383 Articraft assets,
and 264{,}025 assets from \name-400K. PM-only draws exclusively from PM.
Each augmented run first selects PM or its additional source with
probabilities 0.1 and 0.9, then samples uniformly within that source
with replacement. At 150K updates, each run has processed
19.2 million sample draws.
The augmented settings sample PM with probability 0.1,
giving them one tenth of the expected PM exposure
of the PM-only setting.
The comparison fixes the model and number of training updates, while the
additional datasets differ in size and repetition frequency. 

\paragraph{Training and evaluation separation.}
The training subset of \name-400K excludes nine categories: wheelchair, seesaw,
skateboard, telescope, fire hydrant, music stand, wheelbarrow, tricycle,
and walking cane. Semantically corresponding Articraft categories are
excluded before training. Of the 9{,}996 assets in the compiled Articraft
release used here, 225 are removed by this category filter. Preprocessing
and the training/evaluation partition yield the final 9{,}383 training assets.

Within the six PM evaluation categories, we reserve at least 10\% of usable
Articraft objects per category before training, keeping exact rest-geometry
duplicate groups together. This produces 61 evaluation objects. We add 120
additional evaluation objects, contributing 20 objects per category. Their
source identifiers are held out from the training partition. Exact
rest-surface checks compare the evaluation split
internally and against Articraft training geometry. Cross-source near-duplicate
detection and geometric duplicate checks against the complete PM and \name-400K
training sets remain outside the scope of this audit.

\paragraph{In-domain evaluation composition.}
The original PM evaluation split contains 74 objects across six categories,
including 55 storage-furniture objects but only one oven and one washing
machine. We expand this set with 61 Articraft objects reserved before
training and 120 additional objects from the same categories. The resulting
255-object evaluation split is shared by all three models.
Table~\ref{tab:downstream-id-composition} gives the category counts.
The expansion reduces the storage-furniture share from 74.3\% to 33.7\%
and increases the minimum category count to 27. Category and source counts
remain unequal. 

\begin{table}[!htbp]
\centering
\caption{Category and source composition of the in-domain evaluation split.
The added objects belong to the same six categories as the original
PartNet-Mobility evaluation split. All three models are evaluated on the
complete combined split with equal weight per object.}
\label{tab:downstream-id-composition}
\tablefont
\setlength{\tabcolsep}{5pt}
\renewcommand{\arraystretch}{1.12}
\begin{tabular}{@{}lrrrr@{}}
\toprule
\textbf{Category} & \textbf{PM} & \textbf{Articraft} &
\textbf{\name{} (new)} & \textbf{Total} \\
\midrule
Dishwasher & 4 & 4 & 20 & 28 \\
Oven & 1 & 11 & 20 & 32 \\
Refrigerator & 4 & 12 & 20 & 36 \\
Storage furniture & 55 & 11 & 20 & 86 \\
Table & 9 & 17 & 20 & 46 \\
Washing machine & 1 & 6 & 20 & 27 \\
\midrule
\textbf{Total} & \textbf{74} & \textbf{61} & \textbf{120} & \textbf{255} \\
\bottomrule
\end{tabular}
\end{table}

\paragraph{Category-OOD evaluation.}
The category-OOD evaluation contains 774 objects from the nine excluded
categories listed above, with equal weight per object. These categories are
absent from all three training sets. We reuse earlier PM-only and
PM+\name-400K evaluations and evaluate PM+Articraft separately
on another machine.
The checkpoint steps and evaluation protocol match,
but the PM-only and PM+RFW-A-400K settings were not
rerun in the new environment to verify numerical
consistency.

\paragraph{Metrics and aggregation.}
Each inference uses 102{,}400 input points, and metrics use 100{,}000
evaluation points. Rest-pose metrics comprise mean IoU (mIoU), generalized
IoU (gIoU), and part-wise bidirectional Chamfer distance (PC).
Predicted and ground-truth parts
are matched using the public Particulate evaluator. An unmatched part
receives its prescribed penalty: $-1$ for gIoU, $0$ for mIoU, and half
the input bounding-box diagonal for PC. After moving parts to their
maximum articulated states, we report gIoU, PC, and whole-object
bidirectional Chamfer distance (OC). OC compares the complete moved
point clouds without part matching.

For each model, training seed, and evaluation seed, we average per-object metrics over
all 255 in-domain objects or all 774 category-OOD objects, then average over
the five fixed evaluation seeds
$\{10101,20202,30303,40404,50505\}$. Training source probabilities are
not used as evaluation weights. The reported means aggregate three training
seeds $\times$ five evaluation seeds per setting. Statistical significance
within each source or category requires separate evaluation.
Relative gains in Table~\ref{tab:downstream-particulate} are computed
against PM+Articraft from the five-decimal means: $(s-b)/b$ for
higher-is-better metrics and $(b-s)/b$ for distances, where $s$ is the
PM+\name-400K score and $b$ is the PM+Articraft score.

\paragraph{Original PM evaluation split.}
Table~\ref{tab:downstream-pm74} reports a separate evaluation of the
150K checkpoints on the original 74-object PM evaluation split. PM-only has the
best rest-pose mIoU and gIoU and articulated gIoU. PM+Articraft has the
lowest rest-pose PC, while PM+\name-400K has the lowest articulated PC and OC.
PM+\name-400K improves five of six metrics over PM+Articraft; rest-pose PC
is the exception. These scores are not
a source-level breakdown of the combined evaluation results: the two evaluations
use independently prepared ground-truth caches and different sample ordering.

\begin{table}[!htbp]
\centering
\caption{Articulation estimation on the original PartNet-Mobility evaluation
split using the three 150K checkpoints.
Results average three training seeds $\times$ five evaluation seeds; bold marks
the best result. This original split contains 55 storage-furniture objects and
19 objects across the other five categories.}
\label{tab:downstream-pm74}
\tablefont
\setlength{\tabcolsep}{3pt}
\renewcommand{\arraystretch}{1.12}
\begin{tabular}{@{}lcccccc@{}}
\toprule
\multirow{2}{*}{\textbf{Training data}} &
\multicolumn{3}{c}{\textbf{Rest-Pose Segmentation}} &
\multicolumn{3}{c}{\textbf{Articulated Geometry}} \\
\cmidrule(lr){2-4}\cmidrule(lr){5-7}
& mIoU $\uparrow$ & gIoU $\uparrow$ & PC $\downarrow$ &
gIoU $\uparrow$ & PC $\downarrow$ & OC $\downarrow$ \\
\midrule
PM-only & \textbf{0.79658} & \textbf{0.75910} & 0.02554 & \textbf{0.73732} & 0.09812 & 0.01114 \\
PM+Articraft & 0.77070 & 0.73300 & \textbf{0.02276} & 0.70978 & 0.09562 & 0.01232 \\
PM+\name-400K & 0.78706 & 0.74752 & 0.02442 & 0.72360 & \textbf{0.08508} & \textbf{0.01028} \\
\bottomrule
\end{tabular}
\end{table}

\FloatBarrier

\FloatBarrier
\Needspace{8\baselineskip}
\subsection{Generator Component Ablation}
\label{app:generator-ablation}
\label{sec:generator-ablations}

\subsubsection*{Completed Category-Level Results}
\phantomsection
\addcontentsline{toc}{subsubsection}{Completed Category-Level Results}

We evaluate the generator components on \texttt{flip\_phone},
\texttt{microscope}, and \texttt{folding\_chair} under a unified paired
protocol. For each category, all four variants use the same raw source pool,
backbone, execution environment, seed protocol, rendering settings,
evaluators, and failure-handling rules; only the ablated component changes.
Tables~\ref{tab:generator-ablation-summary}
and~\ref{tab:generator-ablation-component-scores} summarize the results.
Mean Judge, component scores, Avg. Iter., and Est. Cost are category means,
while Strict Valid is pooled across all evaluation attempts.

\begin{table}[H]
\caption{Aggregate generator-component ablation outcomes across
\texttt{flip\_phone}, \texttt{microscope}, and \texttt{folding\_chair}.
Mean Judge and Avg. Iter. are category means; Strict Valid reports pooled counts.}
\label{tab:generator-ablation-summary}
\centering
\small
\setlength{\tabcolsep}{7pt}
\renewcommand{\arraystretch}{1.04}
\begin{tabular}{lcccc}
\toprule
\textbf{Variant} & \stackcell{\textbf{Mean}\\\textbf{Judge}} &
\stackcell{\textbf{Strict}\\\textbf{Valid}} &
\stackcell{\textbf{Avg.}\\\textbf{Iter.}} &
\stackcell{\textbf{Est.}\\\textbf{Cost (\$)}} \\
\midrule
w/o Component Extraction & 0.500 & 35/108 & 9.33 & 3.80 \\
w/o Structural Blueprint & 0.401 & 72/108 & 11.33 & 3.73 \\
w/o Validation \& Repair & 0.241 & 0/108 & 0.00 & 0.83 \\
\textbf{\name{} (full)} & \textbf{0.580} &
\textbf{108/108} & \textbf{8.00} & 6.73 \\
\bottomrule
\end{tabular}
\vspace{2pt}
\parbox{\textwidth}{\footnotesize
\emph{Metric definitions.} \textbf{Mean Judge} summarizes geometry, aesthetics,
and semantic quality on a 0--1 scale, with equal weight for each category.
Generation or rendering failures score zero. Scoring details are given in
Table~\ref{tab:generator-ablation-component-scores}.
\textbf{Strict Valid} reports assets passing all required mechanical checks
over total evaluation attempts; failed generation, missing outputs, crashes, and timeouts count as
failures.
\textbf{Avg. Iter.} is the average number of repair iterations per category,
including unsuccessful runs. Zero denotes a disabled repair loop; bold marks
the lowest average among repair-enabled variants.
\textbf{Est. Cost} is the average estimated API cost of building a category
generator, calculated from token usage and the corresponding rates. Failed
runs are included; post-hoc evaluation and figure rendering are excluded.}

\end{table}

\begin{table}[H]
\caption{Mean component scores across \texttt{flip\_phone},
\texttt{microscope}, and \texttt{folding\_chair} for the same ablations.
Each category receives equal weight; higher is better and bold marks the best
mean in each column.}
\label{tab:generator-ablation-component-scores}
\centering
\small
\setlength{\tabcolsep}{7pt}
\renewcommand{\arraystretch}{1.06}
\begin{tabular}{lcccc}
\toprule
\textbf{Variant} & \textbf{Geometry} & \textbf{Aesthetics} &
\textbf{Semantic} & \textbf{Judge} \\
\midrule
w/o Component Extraction & 5.17 & 4.83 & 6.50 & 0.500 \\
w/o Structural Blueprint & 4.67 & 3.83 & 5.33 & 0.401 \\
w/o Validation \& Repair & 2.42 & 2.83 & 3.25 & 0.241 \\
\textbf{\name{} (full)} & \textbf{6.25} & \textbf{5.42} & \textbf{7.00} &
\textbf{0.580} \\
\bottomrule
\end{tabular}
\vspace{2pt}
\parbox{\textwidth}{\footnotesize
\emph{Judge criteria and protocol.} Geometry assesses completeness,
proportions, assembly, and visible geometric defects. Aesthetics assesses
visual coherence, refinement, and design quality. Semantic fidelity assesses
consistency with the target category and the presence of expected functional
parts. For successfully evaluated assets, each dimension is rated from 1 to 10
and normalized as $s=\max(0,\min(1,(v-1)/9))$; per-asset Judge is the mean of
the three normalized scores. The prescribed evaluation protocol requires
four held-out Judge seeds for every category and variant, selected from that
category's hidden seed pool; all four variants share the same four seed IDs.
Rendering settings and the Judge model, model version, prompt, temperature,
and response parser are fixed. Judging is anonymous and blind: method names,
variant identities, and Strict Valid outcomes are withheld from the Judge.
Generation or rendering failures receive zero for all three dimensions and
Judge and remain in the denominator. Scores are averaged within each category
and then across categories with equal weight.}
\end{table}

\begin{figure}[!htbp]
\centering
\includegraphics[width=0.90\textwidth]{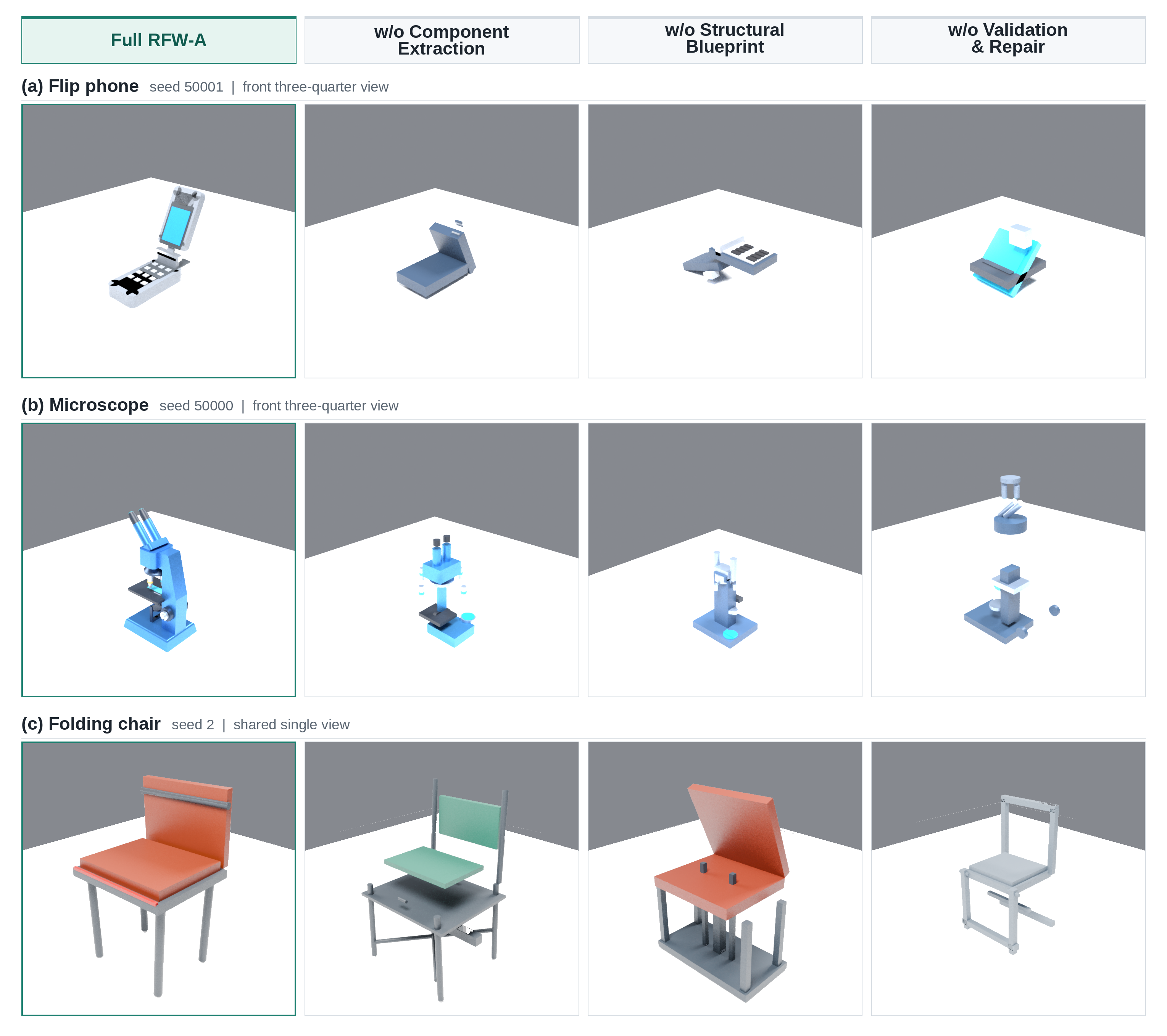}
\caption{\textbf{Qualitative comparison of generator-component ablations.}
Rows show \textbf{(a)} flip phone, \textbf{(b)} microscope, and
\textbf{(c)} folding chair. From left to right, columns compare full \name{}
with variants removing Component Extraction, Structural Blueprint, and
Validation \& Repair, respectively. Quantitative results are reported in
Tables~\ref{tab:generator-ablation-summary}
and~\ref{tab:generator-ablation-component-scores}. The panels are qualitative visual examples; mechanical validity is measured by the strict-validity probes in these tables.}
\label{fig:generator-ablation-combined}
\end{figure}

The ablation separates the roles of the three generator components. The full
configuration reaches the highest mean Judge score (0.580) and passes all 108
strict validity probes. Removing component extraction reduces the score to
0.500 with 35/108 probes passing, and removing the structural blueprint reduces
it to 0.401 with 72/108 passing. Removing validation and repair reduces the
score to 0.241 and yields 0/108 strict-valid assets. The consistent gap from
the full configuration demonstrates the contribution of each component to
quality and executable validity. The full configuration costs \$6.73 per
category on average over the three ablated categories; this estimate reflects
the complete synthesis and repair process.

\Needspace{8\baselineskip}
\subsection{Cabinet Generator Families}
\label{app:cabinet-generators}

\begin{figure}[!htbp]
\centering
\includegraphics[width=0.95\textwidth]{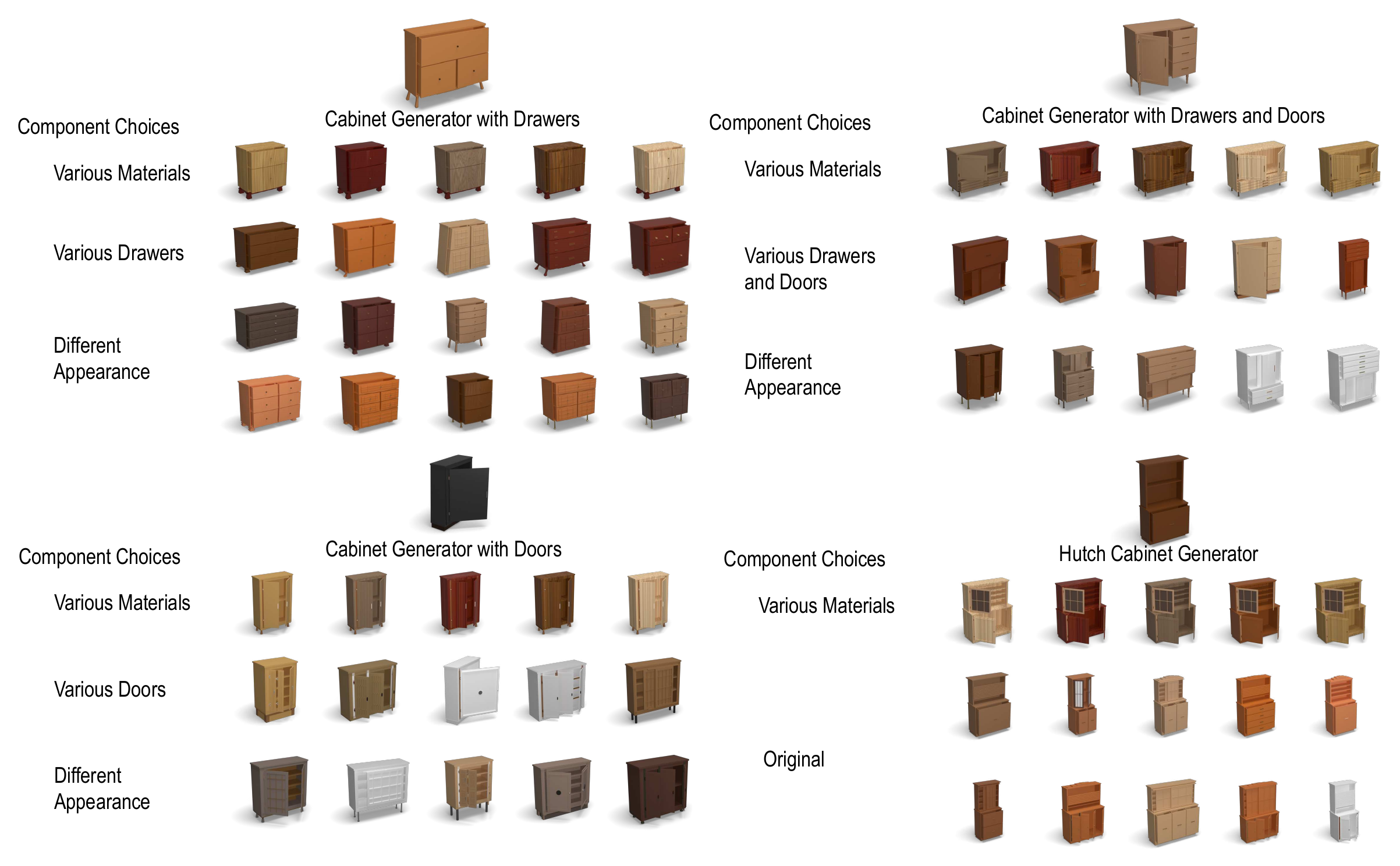}
\caption{\textbf{Cabinet generator families.} Each panel shows assets sampled
from one reusable procedural cabinet generator: \textbf{(top left)} a cabinet
generator with drawers, \textbf{(top right)} a cabinet generator with drawers
and doors, \textbf{(bottom left)} a cabinet generator with doors, and
\textbf{(bottom right)} the hutch cabinet generator of
Figure~\ref{fig:cabinet-generators}, whose row labeled Original reproduces the
reference asset; all other rows and columns are generated variants. Within each panel, rows vary component choices, materials,
drawer and door layouts, and appearance while the generators preserve shared
rules for panel sizing, part placement, and joint attachment.}
\label{fig:cabinet-generator-families}
\end{figure}

Figure~\ref{fig:cabinet-generators} in the main text shows variants drawn from
a single reusable cabinet generator.
Figure~\ref{fig:cabinet-generator-families} extends this example to four
cabinet generators: a cabinet generator with drawers, a cabinet generator with
drawers and doors, a cabinet generator with doors, and the hutch cabinet
generator of Figure~\ref{fig:cabinet-generators}. The four generators encode
different component vocabularies and proportions, so their families differ in
the available drawer and door configurations. Within each family, sampled
assets vary component choices, materials, drawer and door layouts, and
appearance, while panel sizing, part placement, and joint attachment follow
the shared rules encoded once in the generator. Each generator reproduces its
reference configuration and produces structurally consistent variants without
per-asset re-authoring. These families illustrate how resampling a shared program
produces diverse configurations while preserving
category-level construction logic.

\section{Future Work}
\label{app:future}

RFW-A could support the construction of larger-scale embodied datasets and
benchmarks, including synthetic demonstrations, VLA training, and bimanual
or sim-to-real settings that go beyond the five tasks evaluated in
Section~\ref{sec:robotics}
\citep{liu2023libero,wang2024robogen,kim2024openvla,mu2025robotwin,
chen2025robotwin2,zhang2026dexora,cui2026liberosafety}.

The articulated assets and physical annotations in \name-400K could also serve as building
blocks for interactive 3D/4D scene modeling and generative simulation
\citep{chen2026ovow,chen2026lumera,liu2026lightx,xu2025cruise,
li2026genncap,chen2026dar,he2026neoworldpro,wang2026embodiedgenv2,
xia2026sage,pfaff2026scenesmith,qin2026lucida,guo2025dist4d,shuai2026gennav}.

More broadly, RFW-A's executable category-level generators could be integrated
with general agentic 3D generation and editing systems, enabling agents to
construct and modify programmable, articulated, and simulation-ready assets
\citep{lin2026procedura,noor2026nova3d,xia2025editformer,li2025voxhammer,
weng2026feedforward,chi2026vinedresser,ye2026buffalo,nie2026editflow3d}.

\section{Discussion and Limitations}
\label{sec:discussion}

\name allows assets to be resampled and edited without a new agent call,
but diversity remains limited by the source components and category
blueprints. Expanding mechanism and object-family coverage requires extending
these sources or blueprints. The cost benefit depends on construction and repair costs and the
number of valid outputs; broader multi-category studies are needed to
quantify savings and component contributions.

Validation covers sampled configurations and motion states; the full
configuration space remains untested. Collisions can remain despite valid files
(Appendix~\ref{app:structural-integrity}). Real-world dynamic accuracy requires
additional validation, and material properties are estimated under
reference conditions rather than measured per object. Broader motion tests
and validation against real-object behavior remain necessary.

Articulation-estimation results use one model family and three training seeds
per setting. Category-OOD objects are drawn from the released corpus, leaving
transfer to independently sourced assets untested. Dataset size, category
coverage, and asset quality vary jointly under the matched training budgets;
external benchmarks, additional models, and controlled data-size
comparisons are needed.

\end{document}